\documentclass{article}
\PassOptionsToPackage{numbers,sort&compress}{natbib}

\usepackage[final]{neurips_data_2021}

\usepackage[title]{appendix}
\usepackage[T1]{fontenc}
\usepackage{amsfonts}
\usepackage{amsmath}
\usepackage{booktabs}
\usepackage{color}
\usepackage{colortbl}
\usepackage[pdftex]{graphicx}
\usepackage[utf8]{inputenc}
\usepackage{microtype}
\usepackage{multirow}
\usepackage{nicefrac}
\usepackage[olditem,oldenum]{paralist}
\usepackage{soul}
\usepackage{tabularx}
\usepackage{wrapfig}
\usepackage[dvipsnames]{xcolor}

\usepackage[pagebackref,breaklinks,colorlinks,bookmarks=False]{hyperref} 
\usepackage{cleveref}
\usepackage{tikz}

\usepackage{inconsolata}

\hypersetup{
   urlcolor=blue,
   citecolor=ForestGreen,
}

\newcommand{\dsetname}{RedCaps}
\newcommand{\dsetsize}{12M}
\newcommand{\dsetsubs}{350}

\newcommand{\imagenet}[0]{ImageNet-1k}
\newcommand{\voc}[0]{PASCAL VOC}

\newcommand{\sbu}{SBU}
\newcommand{\gcc}{CC-3M}
\newcommand{\Gcc}{CC-12M}

\newcommand{\vnl}{V\&L}
\newcommand{\subreddit}[1]{\texttt{\small \hyperlink{https://reddit.com/r/#1}{\textcolor{blue}{r/#1}}}}
\newcommand{\inlinecap}[1]{\texttt{\textcolor{RoyalBlue}{#1}}}

\newcommand{\virtex}{VirTex}
\newcommand{\virtexii}{VirTex-v2}

\newcolumntype{Z}{>{\centering\arraybackslash}X}

\newcommand{\ttbf}[1]{\textbf{\texttt{#1}}}

\def\checkmark{\tikz\fill[scale=0.4](0,.35) -- (.25,0) -- (1,.7) -- (.25,.15) -- cycle;} 
\newcommand{\band}{\rowcolor{Brown!10}}

\renewcommand{\paragraph}[1]{\vspace{2pt} \noindent \textbf{#1}}

\newcommand{\dsquestion}[1]{
    \item \noindent \textcolor{black}{\textbf{#1}} \smallskip
}
\newcommand{\dsquestionex}[2]{
    \item \noindent \textcolor{black}{\textbf{#1} \textit{#2}} \smallskip
}
\newcommand{\dsanswer}[1]{
    \hspace{-14pt} -- \hspace{10pt}
    \textcolor{Blue}{#1} \medskip
}
\newcommand{\qref}[1]{\textcolor{red}{Q}\ref{#1}}

\title{\dsetname{}: Web-curated image-text data\\created \emph{by the people, for the people}}

\author{
  Karan Desai \quad\quad Gaurav Kaul \quad\quad Zubin Aysola \quad\quad Justin Johnson \\
  University of Michigan \\
  \texttt{\{kdexd,kaulg,aysola,justincj\}@umich.edu} \\
  \url{https://redcaps.xyz} \\
}

\begin{document}

\maketitle

\begin{abstract}
    Large datasets of paired images and text have become increasingly popular for learning generic representations for vision and vision-and-language tasks.
    Such datasets have been built by querying search engines or collecting HTML alt-text
    -- since web data is noisy, they require complex filtering pipelines to maintain quality.
    We explore alternate data sources to collect high quality data with minimal filtering.
    We introduce \dsetname{} -- a large-scale dataset of \dsetsize{} image-text pairs collected from Reddit.
    Images and captions from Reddit depict and describe a wide variety of objects and scenes.
    We collect data from a manually curated set of subreddits, which give coarse image labels and allow us to steer the dataset composition without labeling individual instances.
    We show that captioning models trained on \dsetname{} produce rich and varied captions preferred by humans,
    and learn visual representations that transfer to many downstream tasks.
\end{abstract}

\section{Introduction}
\label{sec:introduction}

\begin{figure}[h]
    \vspace{-10pt}
    \footnotesize
    \setlength \tabcolsep{1pt}
    \newcommand{\greenband}{\rowcolor{ForestGreen!10}}

    \newcolumntype{Y}{>{\centering\arraybackslash}X}
    \begin{tabularx}{\linewidth}{YYYYY}
        \includegraphics[width=0.19\textwidth]{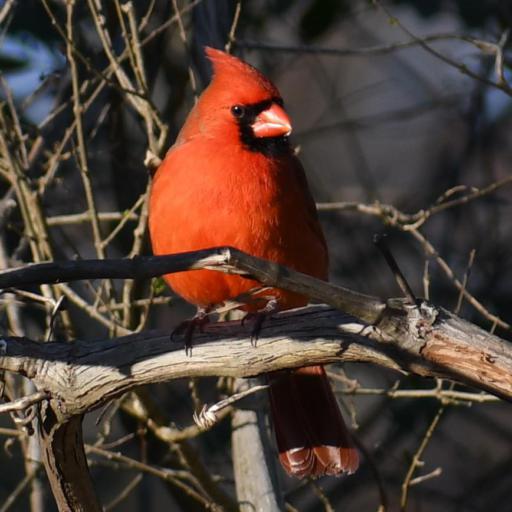} &
        \includegraphics[width=0.19\textwidth]{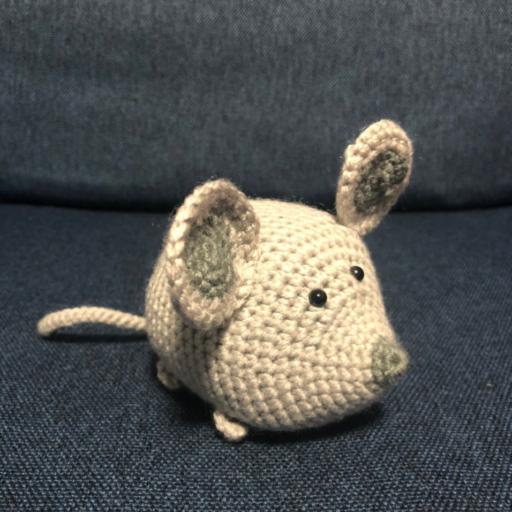} &
        \includegraphics[width=0.19\textwidth]{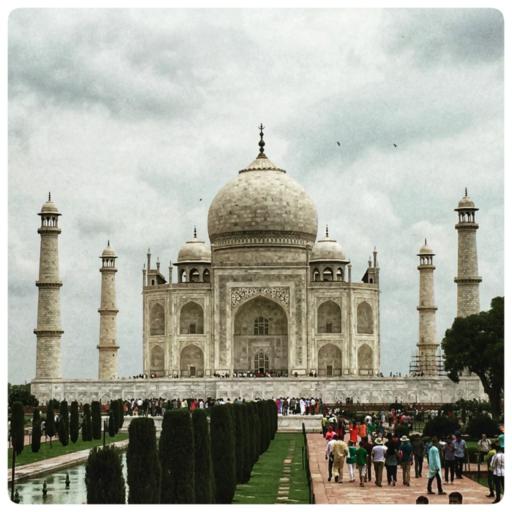} &
        \includegraphics[width=0.19\textwidth]{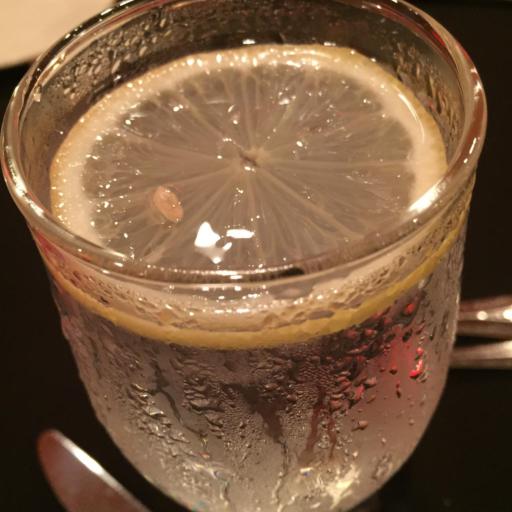} &
        \includegraphics[width=0.19\textwidth]{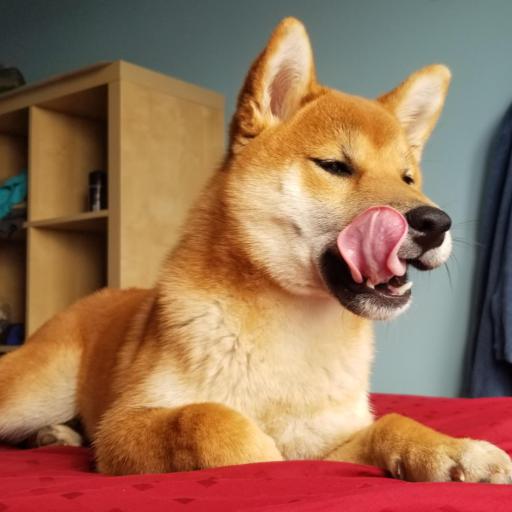}
        \\
        \greenband
        \ttbf{r/birdpics:} male northern cardinal &
        \ttbf{r/crafts:} my mom tied this mouse &
        \ttbf{r/itookapicture:} itap of the taj mahal &
        \ttbf{r/perfectfit:} this lemon in my drink &
        \ttbf{r/shiba:} mlem!
        \\
    \end{tabularx}
    \caption{
        \textbf{\dsetname{} dataset}
        comprises \dsetsize{} image-text pairs from \dsetsubs{} subreddits.
        \dsetname{} data is created \emph{by the people, for the people} -- it contains everyday things that users like to share on social media, for example hobbies (\ttbf{r/crafts}) and pets (\ttbf{r/shiba}).
        Captions often contain specific and fine-grained descriptions (\textcolor{Brown}{northern cardinal}, \textcolor{Brown}{taj mahal}).
        Subreddit names provide relevant image labels (\ttbf{r/shiba}) even when captions may not (\textcolor{Brown}{mlem!}), and sometimes may group many visually unrelated images through a common semantic meaning (\ttbf{r/perfectfit}).
    }
    \label{fig:teaser}
\end{figure}

Large datasets of image-text pairs from the web have enabled successful transfer learning applications in computer vision.
Two such prominent datasets -- SBU~\cite{ordonez2011im2text} and Conceptual Captions~\cite{sharma2018conceptual} --
are widely used for pre-training vision-and-language (\vnl{}) representations~\cite{tan2019lxmert,lu2019vilbert,li2019visualbert,su2019vl,li2020unicoder,chen2019uniter,zhou2020vlp,li2020oscar,huang2020pixelbert}
that transfer to a variety of downstream \vnl{} tasks like visual question answering~\cite{antol2015vqa,zhu2016visual7w,hudson2019gqa}, visual reasoning~\cite{suhr2019corpus,zellers2019recognition}, and image captioning~\cite{chen2015microsoft,agrawal2019nocaps}.
Recent work~\cite{desai2020virtex,bulent2020icmlm} also shows that image-text data from COCO~\cite{chen2015microsoft} can be used to learn \emph{visual} features that are competitive with supervised pretraining~\cite{he2016deep} on ImageNet~\cite{russakovsky2015imagenet,deng2009imagenet} when transfered to downstream tasks~\cite{everingham2009voc,zhou2014places,lin2014microsoft,gupta2019lvis,van2018inaturalist}.
More recently, CLIP~\cite{radford2021clip} and ALIGN~\cite{jia2021scaling} scale up to 400M and 1B+ web-curated image-text pairs, enabling zero-shot visual recognition.

These datasets have an appealing advantage -- they are free from expensive annotations.
However, they apply complex filtering steps to deal with noisy web data.
For example, Conceptual Captions (\gcc{}~\cite{sharma2018conceptual}, \Gcc{}~\cite{changpinyo2021conceptual}) discard captions without nouns, or whose nouns do not match with image labels predicted by in-house image taggers.
They also perform text pre-processing like replacing proper nouns with common nouns.
These pipelines are data-inefficient -- for example, \gcc{} collected 5B image-text pairs and filtered them down to 3.3M.
CLIP and ALIGN scale primarily by \emph{relaxing} such filtering, resulting in gargantuan datasets which could be extremely noisy.

How can we obtain high-quality image-text data from the web \emph{without} complex data filtering?
We argue that the quality of data depends on its \emph{source} and the \emph{intent} behind its creation.
Revisiting data sources, \sbu{} query Flickr with predefined keywords while \gcc{} and \Gcc{} extract images and HTML alt-text from an unspecified set of web pages;
CLIP and ALIGN give only vague descriptions of their data sources, and their datasets are non-public.
In these sources, text is secondary to images:
Flickr focuses on photos, and alt-text is an oft-overlooked \emph{fallback} when images cannot be viewed
that frequently contains metadata or generic text (e.g. ``alt img''~\cite{jia2021scaling}).
To obtain higher-quality data, we look for sources where humans use both images and text equally for interaction on the web.

In this paper, we explore the Reddit~\cite{reddit} social media platform for collecting image-text pairs.
Textual data from Reddit is already used for pre-training massive language models~\cite{volske2017tldr,radford2018gpt1,radford2019gpt2,brown2020gpt3} in NLP.
We collect images and their captions as submitted by Reddit users in topic-specific subreddits.
Our dataset of image captions from Reddit (\dsetname{} in short) consists of \dsetsize{} image-text pairs submitted in \dsetsubs{} subreddits between 2008--2020.
\dsetname{} data is created \emph{by the people, for the people} to engage with the broader community.
\Cref{fig:teaser} shows some examples from \dsetname{} --
the captions are more conversational, humorous, emotional, and generally more diverse than HTML alt-text.

Apart from linguistic diversity, Reddit offers many other advantages.
Subreddits provide additional image labels and group related content --
manually selecting subreddits allows us to steer dataset contents without labeling individual instances.
Reddit's \emph{voting} system gives free and organic quality control: unappealing or spam content is actively \emph{downvoted} by users or removed by moderators.
\dsetname{} is one of the largest public image-text datasets, but it is not \emph{static}:
we plan to release regular updates with newly uploaded Reddit content, allowing \dsetname{} to \emph{grow} over time.

We claim that captions written with the intent of human interaction on Reddit are a better source of data than used in other image-text datasets.
To this end, we follow \virtex~\cite{desai2020virtex} to learn visual representations by training image captioning models from scratch.
We find that human evaluators prefer captioning outputs from models trained on \dsetname{} vs \gcc{}.
We also transfer the learned features to \textbf{eleven} different downstream datasets for tasks including image classification, object detection, instance segmentation, and fine-grained recognition using both fine-tuning and language-based zero-shot classification~\cite{radford2021clip}.
We show that features learned on \dsetname{} outperform those learned on \sbu{} or \gcc{}, demonstrating the utility of our data collection strategy.

\section{\dsetname{}: Collecting image-text pairs from Reddit}
\label{sec:data_collection}

\begin{wrapfigure}{r}{0.45\linewidth}
    \centering
    \vspace{-12pt}
    \includegraphics[width=\linewidth]{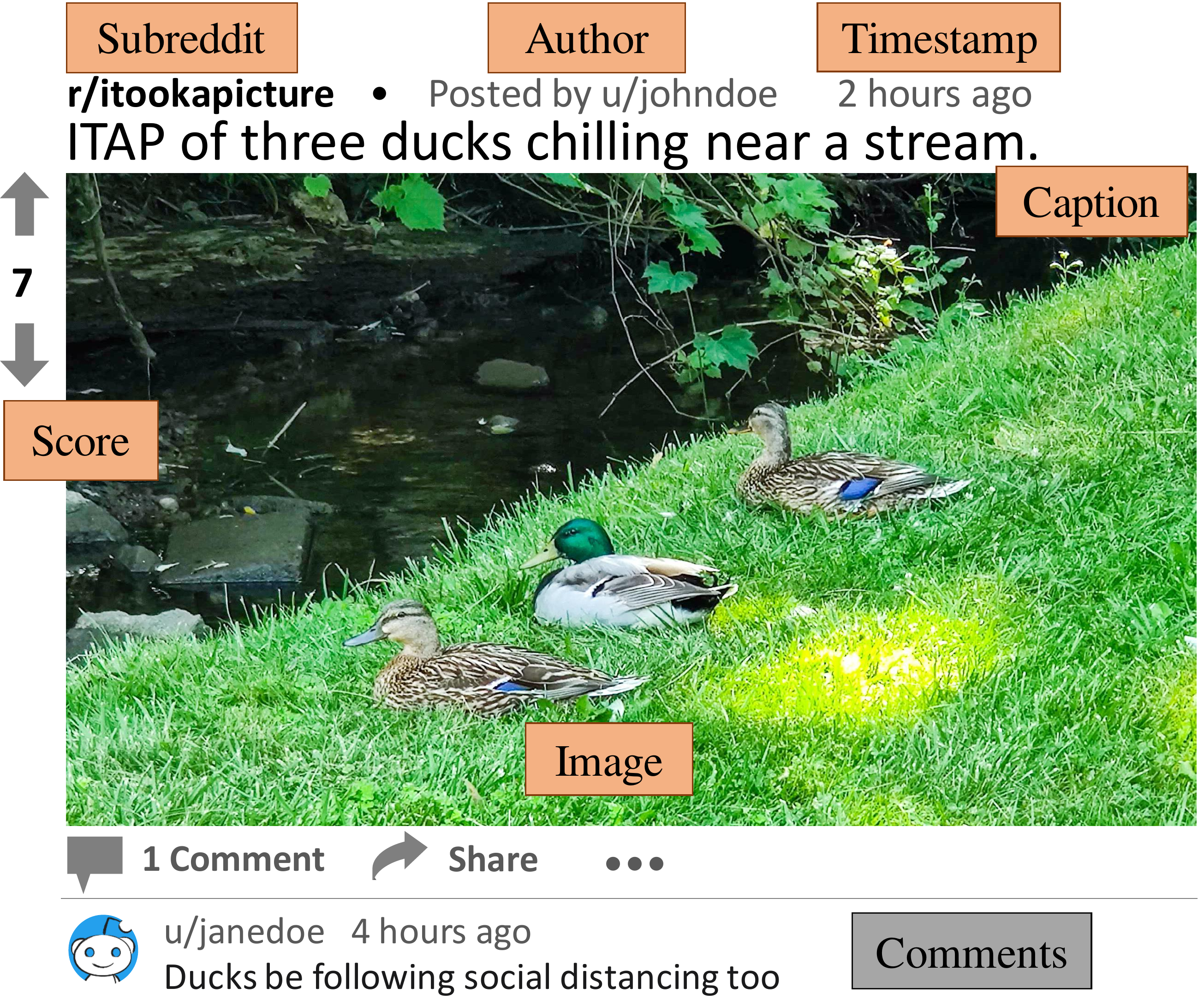}
    \vspace{-12pt}
    \caption{
        \textbf{Preview of a Reddit image post:}
        We build \dsetname{} by extracting images and metadata (in orange) from such image posts.
    }
    \vspace{-36pt}
    \label{fig:image_post}
\end{wrapfigure}

Reddit is the singular data source for \dsetname{}.
This leads to a very different data collection pipeline than datasets based on HTML alt-text or search engine results.
Here we describe how we collect \dsetname{}.

\paragraph{Overview of Reddit:}
Reddit is a social media platform for content sharing and discussion.
It comprises user-run communities called \emph{subreddits} that cover diverse topics like
animals (\subreddit{cats}, \subreddit{foxes}),
food (\subreddit{pizza}, \subreddit{sushi}),
leisure (\subreddit{hiking}, \subreddit{crafts}),
and utility (\subreddit{ceramics}, \subreddit{tools}).
Users can submit new posts or share existing posts from other subreddits (\emph{cross-posting}), and may comment and upvote (or downvote) posts to express their interest.

We are specifically interested in posts containing images.
\Cref{fig:image_post} shows an image post submitted by user \textcolor{blue}{\texttt{u/johndoe}} in subreddit \subreddit{itookapicture}.
It comprises an image, caption, score (upvotes minus downvotes), and information about the author and time of post creation.
We extract this metadata from millions of image posts to build \dsetname{}.

Reddit posts also have associated comment threads.
These are usually casual conversations \emph{loosely} based on the image.
In \Cref{fig:image_post}, the comment describes ducks as following \emph{social distancing} --
it includes context beyond the image (COVID-19 pandemic) and conveys it with a witty remark.
Prior works in dialog modeling and text summarization have
trained on Reddit comments~\cite{rfou2016conversational,dodge2016evaluating,mazare2018training,henderson2019repository,volske2017tldr}.
For \dsetname{}, we only use captions as textual data and leave comments for future work.

\subsection{Data collection pipeline}
\label{subsec:pipeline}

Reddit's uniform structure allows us to parallelize data collection as independent tasks --
each task involves collecting posts submitted to a single subreddit in one year.
Our collection pipeline has three steps:
(1) subreddit selection,
(2) image post filtering, and
(3) caption cleaning.

\paragraph{Step 1. Subreddit selection:}
We collect data from a manually curated set of subreddits.
Subreddits have their own rules, community norms, and moderators so curating subreddits allows us to steer the dataset's composition without annotating individual instances.
We select subreddits with a high volume of images posts, where images tend to be photographs (rather than memes, drawings, screenshots, etc) and post titles tend to describe image content (rather than making jokes, political commentary, etc).
We do not select any NSFW, banned, or quarantined subreddits.
We want to minimize the number of \emph{people} that appear in \dsetname, so we omit subreddits whose primary purpose is to share or comment on images of people (such as celebrity pics or user selfies).
We choose subreddits focused on general photography (\subreddit{pics}, \subreddit{itookapicture}),
animals (\subreddit{axolotls}, \subreddit{birdsofprey}, \subreddit{dachshund}), plants (\subreddit{roses}, \subreddit{succulents}),
objects (\subreddit{classiccars}, \subreddit{trains}, \subreddit{mechanicalkeyboards}), food (\subreddit{steak}, \subreddit{macarons}),
scenery (\subreddit{cityporn}\footnote{Many subreddits are jokingly titled \emph{-porn} to indicate beautiful non-pornographic images.},
\subreddit{desertporn}), or activities (\subreddit{carpentry}, \subreddit{kayaking}).
In total we collect data from \dsetsubs{} subreddits; the full list can be found in Appendix \ref{appendix:subreddits}.

\paragraph{Step 2. Image post filtering:}
We use Pushshift~\cite{baumgartner2020pushshift} and Reddit~\cite{redditapi,redditpraw} APIs to download all image posts submitted to our selected subreddits from 2008--2020.
Posts are collected at least six months after their creation to let upvotes stabilize.
We only collect posts with images hosted on three domains:
Reddit ({\small \url{i.redd.it}}), Imgur ({\small \url{i.imgur.com}}), and Flickr ({\small \url{staticflickr.com}}).
Some image posts contain multiple images (\emph{gallery posts}) -- in this case we only collect the first image and associate it with the caption.
We discard posts with $< 2$ upvotes to avoid unappealing content,
and we discard posts marked NSFW (by their authors or subreddit moderators) to avoid pornographic or disturbing content.

\paragraph{Step 3. Caption cleaning:}
We expect Reddit post titles to be less noisy than other large-scale sources of image captions such as alt-text~\cite{sharma2018conceptual,changpinyo2021conceptual},
so we apply minimal text cleaning.
We lowercase captions and use \texttt{ftfy}~\cite{speer2019ftfy} to remove character accents, emojis, and non-latin characters, following \cite{radford2019gpt2,brown2020gpt3,radford2021clip}.
Then we apply simple pattern matching to discard all sub-strings enclosed in brackets (\inlinecap{(.*)}, \inlinecap{[.*]}).
These sub-strings usually give non-semantic information:
\emph{original content} tags \inlinecap{[oc]},
image resolutions \inlinecap{(800x600 px)},
camera specs \inlinecap{(shot with iPhone)},
self-promotion \inlinecap{[Instagram: @user]},
and other references \inlinecap{(link in comments)}.
Finally, like \cite{changpinyo2021conceptual} we replace social media handles (words starting with `@') with a \inlinecap{[USR]} token to protect user privacy and reduce redundancy.
Due to such filtering, ${\approx}$12K (0.1\%) captions in our dataset are empty strings.
We do not discard them, as subreddit names alone provide meaningful supervision.
Unlike \gcc{} or \Gcc{} that discard captions without nouns or that don't overlap image tags, we do not discard any instances in this step.

Through this pipeline, we collect 13.4M instances from \dsetsubs{} subreddits.
Our collection pipeline is less resource-intensive than existing datasets -- we do not require webpage crawlers, search engines, or large databases of indexed webpages.
\dsetname{} is easily extensible in the future by selecting more subreddits and collecting posts from future years.
Next, we perform additional filtering to mitigate user privacy risks and harmful stereotypes in \dsetname{}, resulting in final size of \dsetsize{} instances.

\vspace{-4pt}
\subsection{Ethical considerations}
\label{subsec:ethics}

There has been growing awareness about potential biases and harms that can arise from internet-scale image and text datasets~\cite{gebru2018datasheets,de2019does,jo2020lessons,paullada2020data,birhane2021wacv,yang2021study,bender2021dangers}.
There is a fundamental tension in such datasets:
the use of internet data is motivated by the desire to use datasets larger than can be manually annotated or verified,
but this also means that such datasets cannot be fully controlled or curated by their creators.

We identify two potential risks with \dsetname{} -- privacy of people appearing in \dsetname{} images, and harmful stereotypes -- and attempt to minimize them by \emph{automatic data filtering}.
We also discuss the impact of data curation from Reddit on user consent and data distribution in \dsetname{}.

\begin{wraptable}{r}{0.5\linewidth}
    \centering
    \small
    \setlength{\tabcolsep}{2pt}
    \renewcommand{\arraystretch}{0.95}
    \vspace{-10pt}
    \begin{tabularx}{\linewidth}{X ccccccc} 
        \toprule
        & \textbf{Detected} && \multicolumn{2}{c}{\textbf{Precision}} && \multicolumn{2}{c}{\textbf{Missed dets.}} \\
        \cmidrule{4-5} \cmidrule{7-8}
        & \textbf{(Filtered)} && \textbf{5K} & \textbf{(\%)} && \textbf{50K} & \textbf{12M} \\
        \midrule
        Faces               & 1.2M && 1615 & 32\% && 79 & $\approx$19K \\
        NSFW                & 87K  && 65 & 1\% && 1 & $\approx$240 \\
        Language$~\dagger$  & 24K  && -- & -- && -- & -- \\
        \bottomrule
    \end{tabularx}
    \caption{
        \textbf{Automatic filtering:} We use detectors to filter $\sim$1.4M instances with images containing faces or NSFW content, or captions containing potentially derogatory language.
        We estimate the \emph{precision} of these detectors by reviewing 5K random detected images.
        After filtering, we review 50K random images (out of 12M) to estimate \emph{missed detections} -- faces and NSFW images remaining in \dsetname{} -- which we find to be extremely low.
    }
    \label{tab:filtering}
    \emph{\footnotesize{$\dagger$: Language filtering is deterministic (string matching).}}
    \vspace{-15pt}
\end{wraptable} 

\paragraph{Privacy:}
The individual who \emph{posts} a given photo on Reddit may not be the person \emph{appearing} in said photo;
this can pose privacy risks for people who did not expect to appear in images online~\cite{birhane2021wacv, yang2021study}.
Our first method of mitigation is the manual curation of subreddits which are not focused on describing people (Section~\ref{subsec:pipeline}).
As an additional measure, we use RetinaFace~\cite{deng2019retinaface} to filter images having any face detection with confidence $\geq 0.9$.
Results of this filtering are shown in \Cref{tab:filtering}.
The number of detections are high (1.2M), however the precision is low (32\%) -- most detections are masked faces, statues, and animals.
Nevertheless we remove all of these images to reduce privacy risks while minimizing impact to downstream vision tasks.

\paragraph{Harmful Stereotypes:}
Another concern with Reddit data is that images or language may represent harmful stereotypes about gender, race, or other characteristics of people~\cite{paullada2020data,birhane2021wacv,bender2021dangers}.
We select only non-NSFW subreddits with active moderation for collecting data.
This stands in contrast to less curated uses of Reddit data, such as GPT-2~\cite{radford2019gpt2} whose training data includes at least 63K documents from banned or quarantined subreddits which may contain toxic language~\cite{gehman2020realtoxicityprompts}.
We attempt to further reduce harmful stereotypes in two ways:
\begin{compactitem}[\hspace{1pt}--]
    \item \textbf{NSFW images:}
    We use the InceptionV3~\cite{szegedy2016rethinking} model from \cite{nsfwinception} to filter images detected as \emph{porn} or \emph{hentai} with confidence $\geq 0.9$.
    Similar to face filtering, we estimated precision of our filtering and estimated amount of missed detections, shown in \Cref{tab:filtering}.
    The model detects 87K images with low precision ($\sim$1\%) -- most detections are non-NSFW images with pink and beige hues.
    \item \textbf{Potentially derogatory language:}
    We filter instances whose captions contain words or phrases from a common blocklist~\cite{ldnoobw}.
    It is important to note that such coarse filtering might suppress language from marginalized groups reclaiming slurs~\cite{bender2021dangers};
    however, as \dsetname{} is not intended to describe people, we believe this is a pragmatic tradeoff to avoid propagating harmful labels.
\end{compactitem}

\paragraph{Consent:}
When submitting to Reddit, users expect their posts to be publicly visible and accessible via the Reddit API we use to download data.
However, they did not explicitly consent for their data to be used for training large-scale neural networks~\cite{birhane2021wacv}.
We mitigate this concern in two ways.
First, we distribute URLs instead of images;
posts deleted from Reddit will thus be automatically removed from \dsetname{}.
Second, we provide a public form allowing anyone to request that specific instances be removed from \dsetname{} on our website.
These decisions mean that over time some image will disappear from \dsetname{},
making it difficult to \emph{exactly} reproduce experiments in the future.
However we believe this to be less important than allowing users to opt out from \dsetname.
Even if images are removed, we expect \dsetname{} to \emph{grow} over time as we include newer posts (\Cref{fig:dataset_size}).

\paragraph{Reddit demographics:}
Reddit's user demographics are not representative of the population at large.
Compared to US adults, Reddit users skew male (69\% vs 49\%), young (58\% 18-29 years old vs 22\%), college educated (36\% vs 28\%), and politically liberal (41\% vs 25\%)~\cite{redditstats}.
Reddit users are predominantly white (63\%)~\cite{redditstats}, and 49\% of desktop traffic to Reddit comes from the United States~\cite{reddittraffic}.
All of the subreddits in \dsetname{} use English as their primary language.
Taken together, these demographic biases likely also bias the types of objects and places that appear in images on Reddit, and the language used to describe these images.
We do not offer explicit countermeasures to these biases, but users of \dsetname{} should keep in mind that \emph{size doesn't guarantee diversity}~\cite{bender2021dangers}.

Subtler issues may also exist, such as imbalanced representation of demographic groups~\cite{buolamwini2018gender} or gender bias in object co-occurrence~\cite{zhao2017men} or language~\cite{hendricks2018women}.
These are hard to control in internet data, so we release \dsetname{} with explicit instructions on suitable use-cases; specifically requesting models not be trained to identify people, or make decisions that impact people.
We document these instructions and other terms-of-use in a datasheet~\cite{gebru2018datasheets}, provided in Appendix \ref{appendix:datasheet}.

\section{\dsetname{} data analysis}
\label{sec:data_analysis}

\begin{figure}[h]
    \centering
    \footnotesize
    \begin{minipage}[t]{0.46\linewidth}
    \vspace{-10pt}
    \includegraphics[width=\linewidth]{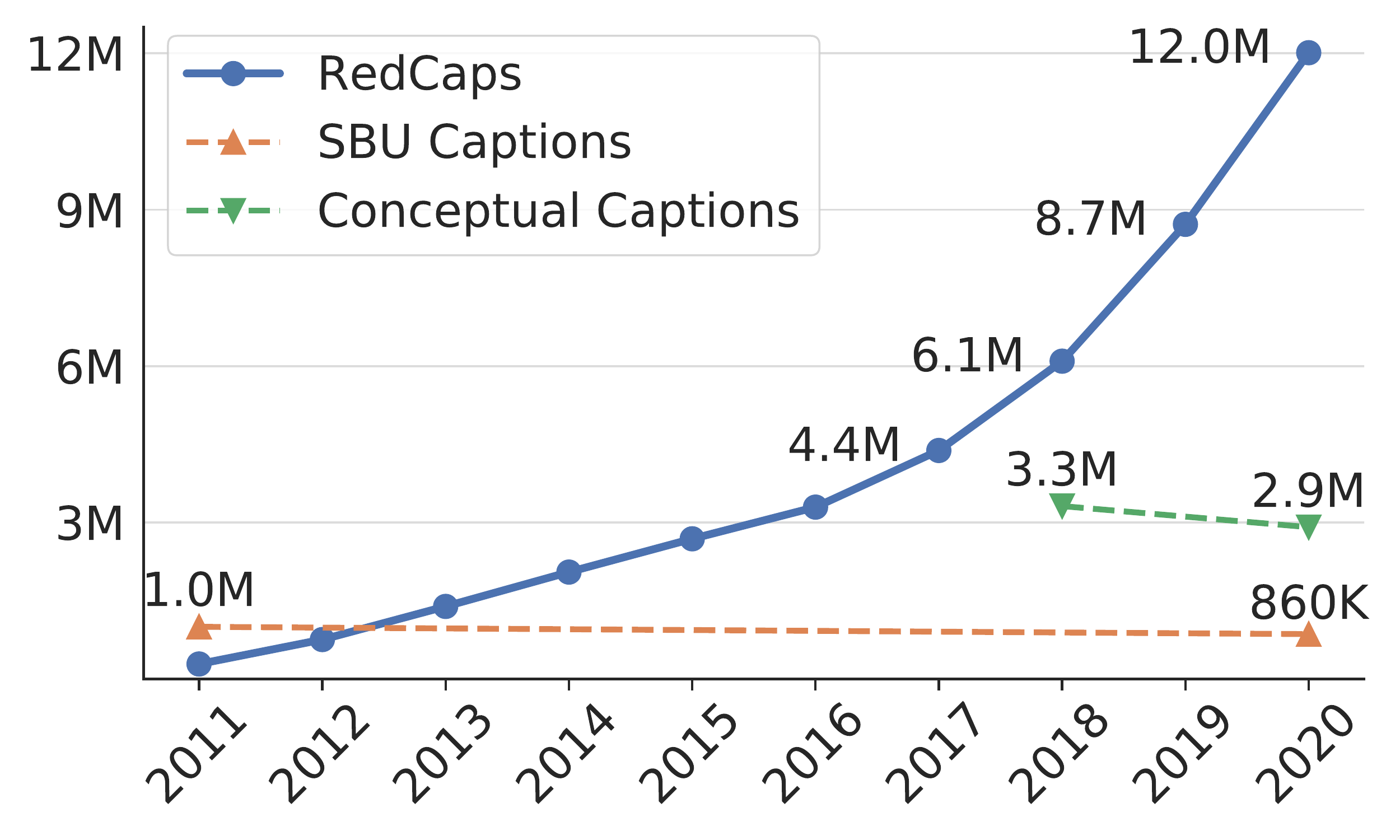} 
    \begin{tabularx}{\linewidth}{lcrr}
        \toprule
        \textbf{Datasets in 2021} && \textbf{\# Instances} & \textbf{Released} \\
        \midrule
        \dsetname{} (ours) && 12,011,111 & \checkmark \\
        \Gcc{}~\cite{changpinyo2021conceptual} && 12,423,374 & \checkmark \\
        WIT-english~\cite{srinivasan2021wit} && 5,500,746 & \checkmark \\
        CLIP~\cite{radford2021clip} && 400M & $\times$ \\
        ALIGN~\cite{jia2021scaling} && 1.8B & $\times$ \\
        \bottomrule
    \end{tabularx}
    \caption{
        \textbf{Dataset size comparison:}
        \dsetname{} is one of the largest public image-text datasets, and is expected to \emph{grow} over time.
    }
    \label{fig:dataset_size}
    \end{minipage}
    \hfill
    \begin{minipage}[t]{0.48\linewidth}
    \vspace{-20pt}
    \includegraphics[width=\linewidth]{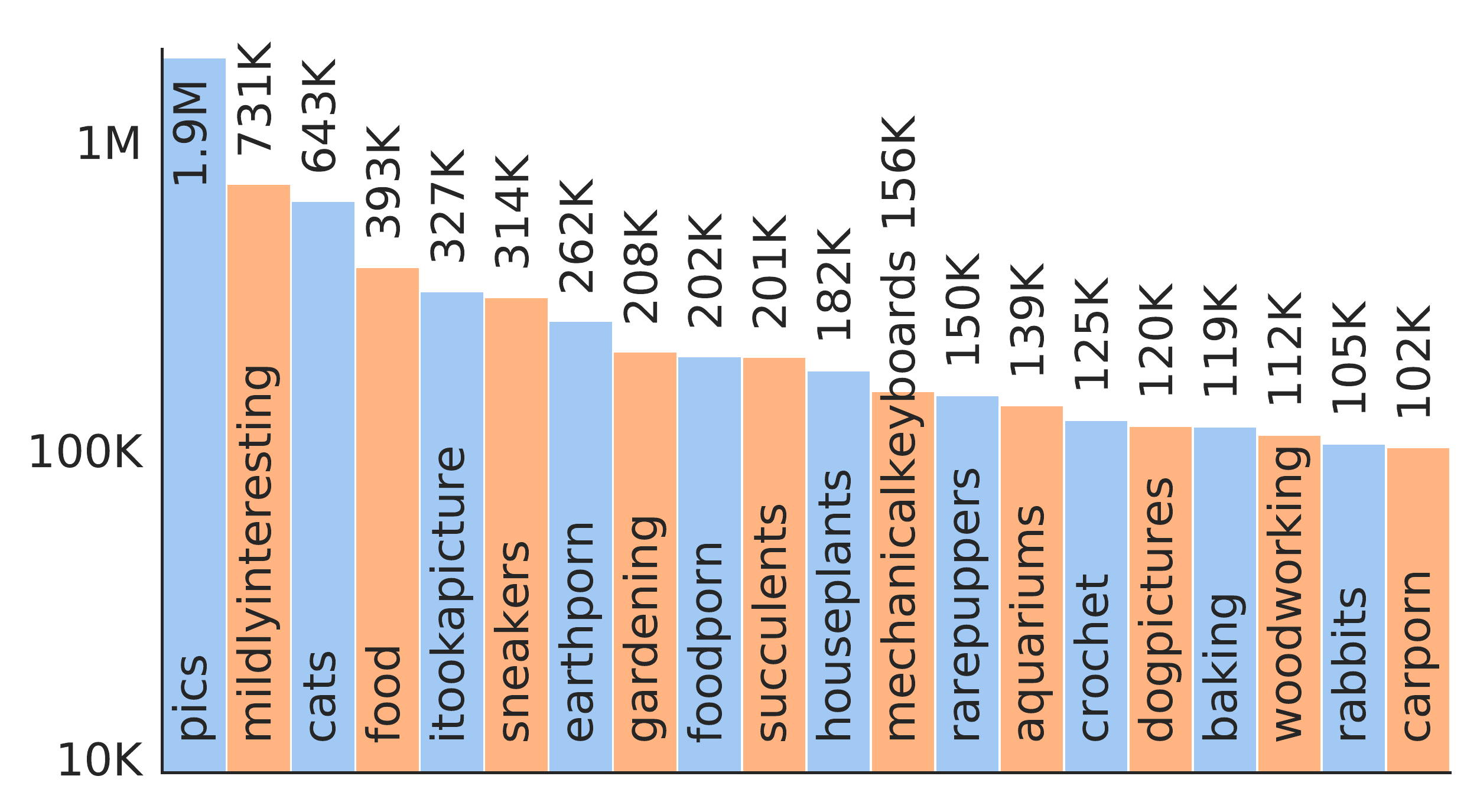}
    \vspace{-20pt}
    \caption{
        \textbf{Instances per subreddit:}
        Top 20 subreddits with most image-text pairs in \dsetname{}.
    }
    \label{fig:subreddit_distribution}
    \vspace{10pt}
    \includegraphics[width=\linewidth]{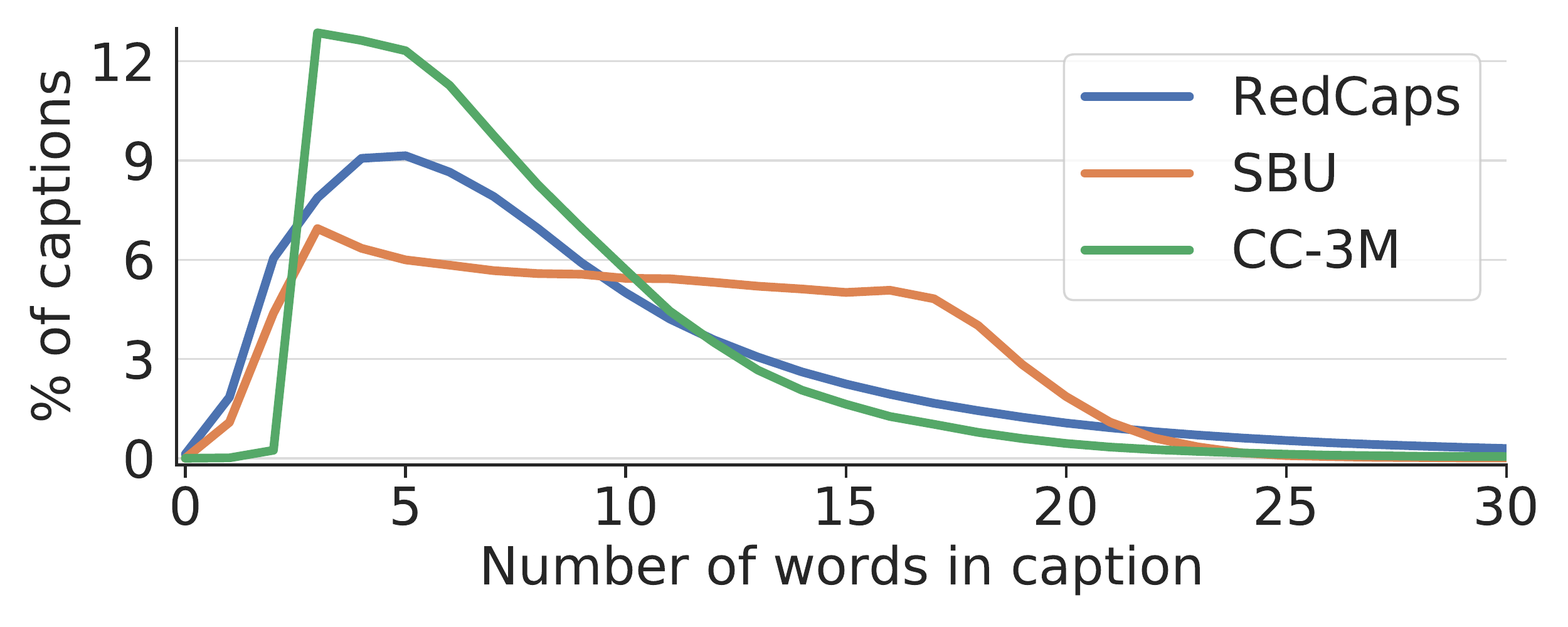}
    \vspace{-20pt}
    \caption{
        \textbf{Caption Lengths:}
        \dsetname{} has a long tailed distribution of caption lengths.
    }
    \label{fig:caption_lengths}
    \end{minipage}
\end{figure}

\paragraph{Dataset size:}
\Cref{fig:dataset_size} \textcolor{Red}{(top)} shows the growth of \dsetname{} between 2011--2020 based on creation timestamps of image posts (see \Cref{fig:image_post}).
We observe that both \sbu{} and \gcc{} have shrunk in size since their release.
Since these datasets have released images as URLs (similar to us), an instance would become invalid if the underlying image is removed from the URL~\footnote{We use full \sbu{} and \gcc{} annotations for analysis instead of discarding captions with invalid URLs.}.
Likewise, some instances in \dsetname{} can also disappear in the future if Reddit users delete their posts.
However, new image posts on Reddit outnumber deleted posts -- we expect \dsetname{} size to increase in future versions.

\Cref{fig:dataset_size} \textcolor{Red}{(bottom)}, compares \dsetname{} with recent image-text datasets released in 2021.
\dsetname{} is 2$\times$ larger than the English subset of multilingual Wikipedia image-text dataset~\cite{srinivasan2021wit}, and nearly as large as \Gcc{}~\cite{changpinyo2021conceptual}.
Based on current trends, we expect \dsetname{} to outsize \Gcc{} by the end of 2021.
While CLIP~\cite{radford2021clip} and ALIGN~\cite{jia2021scaling} used orders of magnitude larger training datasets, they are not released for public use -- \dsetname{} remains one of the largest public image-text datasets.

\paragraph{Subreddit distribution:}
\dsetname{} instances are distributed across \dsetsubs{} subreddits in a long-tail distribution.
In \Cref{fig:subreddit_distribution}, we show top 20 subreddits with most instances in \dsetname{}.
Subreddit sizes highly correlate with their popularity on Reddit, which depends on what users find interesting to view and share on social media.
Large subreddits are based on general photography (\subreddit{pics}, \subreddit{mildlyinteresting}, \subreddit{itookapicture}),
while specific subreddits show that Reddit users enjoy sharing images of food (\subreddit{food}, \subreddit{foodporn}),
cute pets (\subreddit{cats}, \subreddit{dogpictures}, \subreddit{rabbits}),
and show off their hobbies (\subreddit{gardening}, \subreddit{crochet}, \subreddit{baking})
and accesories (\subreddit{sneakers}, \subreddit{mechanicalkeyboards}, \subreddit{carporn}).
This gives a distribution of visual concepts encountered by humans in daily life without having to predefine an ontology of object classes.

\paragraph{Caption lengths:}
\Cref{fig:caption_lengths} compares caption lengths between \dsetname{} and other datasets.
We see that \dsetname{} has the highest mode length at 5 words (vs 3 for \gcc{}, \sbu{})
and a heavier tail of long captions $\geq$25 words.
\sbu{} has a fairly flat distribution of captions between 3 and 17 words, likely since they only retain captions with at least one preposition and two words in a manually curated term list;
\dsetname{} and \gcc{} captions are not filtered in this way and have more peaked distributions reflecting natural language usage.

\begin{figure}[t]
    \centering
    \footnotesize
    \begin{minipage}[t]{0.48\linewidth}
    \setlength\tabcolsep{5pt}
    \begin{tabularx}{\linewidth}{Xrrr}
        \toprule
        \textbf{Dataset} & \textbf{Unigrams} & \textbf{Bigrams} & \textbf{Trigrams} \\
        \midrule
        \sbu{}      & 28,989 & 107,847 & 99,687 \\
        \gcc{}      & 21,223 & 230,077 & 287,017 \\
        \dsetname{} & 95,777 & 770,100 & 866,243 \\
        \bottomrule
    \end{tabularx}
    \begin{tabularx}{\linewidth}{Xr}
        \\
        \multicolumn{2}{c}{\textbf{Top-5 frequent Trigrams}} \\
        \toprule
        \sbu{}          & in front of, black and white, in the sky \\
                        & in the background, in the water \\
        \midrule
                        & a white background, on a white, \\
        \gcc{}          & image may contain, illustration of a \\
                        & may contain person \\
        \midrule
        \dsetname{}     & itap of a, i don't, one of my \\
                        & itap of the, this is my \\
        \bottomrule
    \end{tabularx}
    \makeatletter
    \def\@captype{table}
    \makeatother
    \vspace{-2pt}
    \caption{
        \textbf{Word count statistics:}
        Number of $\{1,2,3\}$-grams occurring at least 10 times \textbf{(top)} and top-5 trigrams in each dataset \textbf{(bottom)}.
    }
    \label{tab:word_counts}
    \end{minipage}
    \hfill
    \begin{minipage}[t]{0.48\linewidth}
    \setlength\tabcolsep{3pt}
    \begin{tabularx}{\linewidth}{lrrrr}
        \toprule
        \textbf{Dataset} & \textbf{C. Nouns} & \textbf{P. Nouns} & \textbf{Adjectives} & \textbf{Verbs} \\
        \midrule
        \sbu{}      & 12,985 & 8,748    &  2,929 & 2,497 \\
        \gcc{}      & 8,116  & 654      &  4,676 & 3,467 \\
        \dsetname{} & 26,060 & 38,405   & 11,029 & 6,019 \\
        \bottomrule
    \end{tabularx}
    \includegraphics[width=\linewidth]{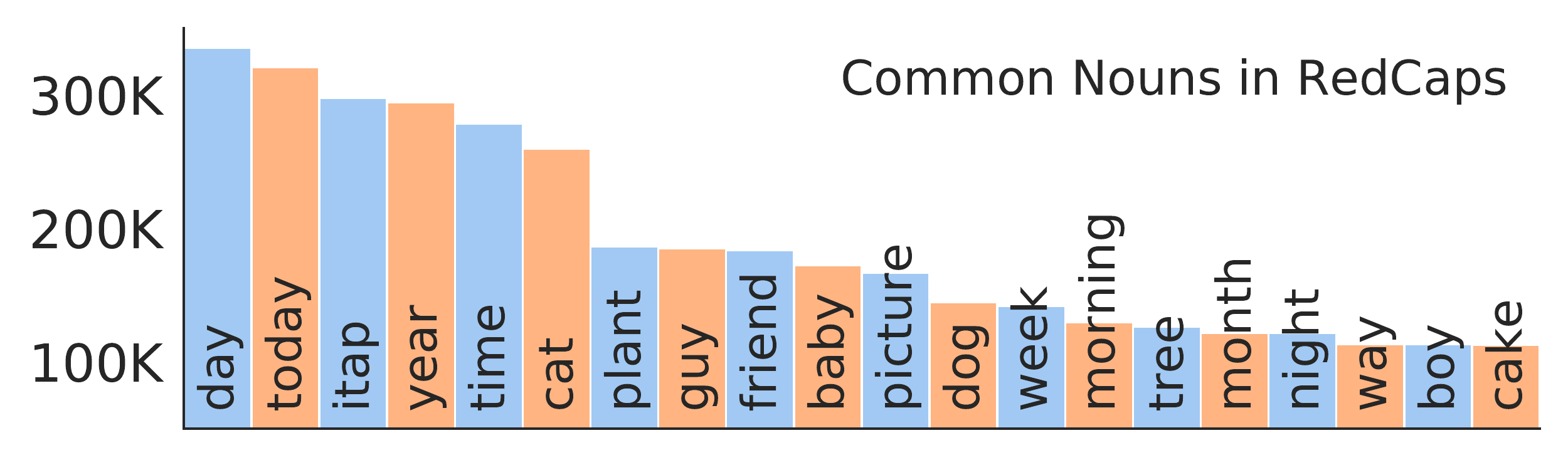}
    \includegraphics[width=\linewidth]{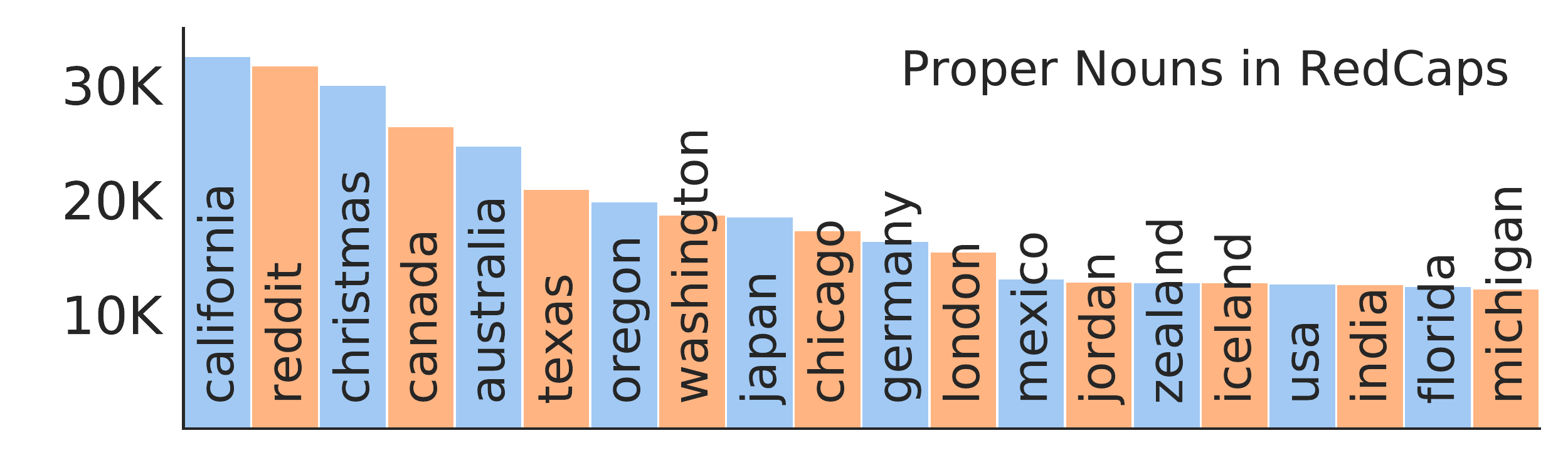}
    \vspace{-20pt}
    \caption{
        \textbf{Linguistic statistics:}
        Number of unique words by POS, occurring at least 10 times \textbf{(top)}, and frequent nouns in \dsetname{} \textbf{(bottom)}.
    }
    \label{fig:linguistic_stats}
    \end{minipage}
\end{figure}

\paragraph{Word count statistics:}
\Cref{tab:word_counts} \textcolor{Red}{(top)} compares linguistic diversity between datasets by computing the number of unique unigrams (words), bigrams, and trigrams occurring at least 10 times.
This reveals that \gcc{} has surprisingly little linguistic diversity, having less unique unigrams than \sbu{} despite having ${\approx}3\times$ \emph{more} captions.
\dsetname{} has the most unique terms, with more than 4$\times$ unigrams and more than 3$\times$ bigrams and trigrams than \gcc{}.
Greater linguistic diversity means that models trained on \dsetname{} should recognize a larger variety of visual concepts.

\Cref{tab:word_counts} \textcolor{Red}{(bottom)} shows the most frequent trigrams per dataset.
\sbu{} has many prepositional phrases, likely since they require all captions to contain a preposition.
Common \gcc{} trigrams \emph{image may contain}, \emph{may contain person} suggest that the alt-text from which \gcc{} takes captions may sometimes be automatically generated.
\dsetname{} trigrams \emph{I don't}, \emph{one of my}, \emph{this is my} are more conversational and draw a personal connection between the author and the image, whereas other trigrams \emph{itap of a} and \emph{itap of the} reflect community conventions on \subreddit{itookapicture}.

\paragraph{Linguistic statistics:}
We use part-of-speech (POS) tagging to dig deeper into linguistic diversity of \dsetname{}.
We use the \texttt{en\_core\_web\_trf} model from SpaCy~\cite{spacy} to tag POS in all captions.
\Cref{fig:linguistic_stats} \textcolor{Red}{(top)} shows number of unique words per POS appearing at least 10 times.
\dsetname{} has ${>}2\times$ more common nouns and ${>}4\times$ more proper nouns than \sbu{}, and ${>}2\times$ more adjectives and ${>}1.5\times$ more verbs than \gcc{}.
Nouns in \gcc{} are artifically deflated, since their pipeline replaces proper nouns and named entities with hypernyms (which may explain their low unigram counts in \Cref{tab:word_counts}).

\Cref{fig:linguistic_stats} \textcolor{Red}{(bottom)} shows the most frequent occurring nouns in \dsetname{}.
We see a variety of common nouns, both concrete (\emph{cat}, \emph{plant}) and abstract (\emph{day}, \emph{time}).
We find that nouns like \emph{guy}, \emph{baby}, and \emph{boy} are frequent with \dsetname{} images with pet animals.
Moreover, most frequent proper nouns comprise many cities (\emph{chicago}, \emph{london}), states (\emph{california}, \emph{texas}), and countries (\emph{japan}, \emph{germany}, \emph{india}), indicating the geographical diversity of \dsetname{}.

\section{Experiments}
\label{sec:experiments}

We aim to show that \dsetname{} offers a unique style of data for both vision and \vnl{} applications.
We demonstrate both applications by adapting \virtex~\cite{desai2020virtex},
a recent method for pre-training visual representations by performing image captioning as proxy task.
In this section, we measure the effect of data quality on downstream vision tasks by training \virtex{} models with the same architecture but different datasets -- \sbu{}, \gcc{}, and \dsetname{}.
To control for \dsetname{}'s size, we also train on a subset of \dsetname{} instances from 2020 -- this has size comparable to \gcc{} (\emph{3.2M vs 2.9M}).

\paragraph{Extending \virtex{} to \virtexii{}:}
\virtex{} comprises an image encoder (\emph{visual backbone}) and a pair of text decoders (\emph{textual head}) that predict the caption token-by-token in forward and backward directions.
The base model from \cite{desai2020virtex} used a ResNet-50~\cite{he2016deep} visual backbone, and Transformers~\cite{vaswani2017attention} in textual head that are $L = 1$ layers deep and $H = 2048$ dimensions wide, and was trained on COCO Captions~\cite{chen2015microsoft} (118K images).
We modify this model from \cite{desai2020virtex} to \virtexii{} in order to scale to larger noisy datasets,
making the following changes:
\begin{compactitem}[\hspace{1pt}--]
    \item \textbf{Model architecture:}
    We use deeper Transformers with $L = 6$ layers.
    To balance the memory requirements, we reduce the width to $H = 512$.
    We use the recent \emph{Pre-LN} Transformer variant~\cite{baevski2019adaptive,wang2019learning,radford2019gpt2} that is more stable to train large transformers~\cite{xiong2020layer} --
    LayerNorm~\cite{ba2016layer} is moved inside the residual connection, and we add LayerNorm before the prediction layer.
    \item \textbf{Tokenization:}
    Similar to \virtex{}, we use SentencePiece tokenizer~\cite{kudo2018sentencepiece} with BPE~\cite{sennrich2016bpe}.
    We build a vocabulary of 30K tokens from the combined caption corpus of \sbu{}, \gcc{} and \dsetname{}.
    For fair comparison, we use the same vocabulary for all models trained on different datasets.
    When training with \dsetname{}, we \emph{prefix} the caption with subreddit tokens: e.g. for \Cref{fig:teaser} (\ttbf{r/birdpics}), the caption becomes \inlinecap{\small \texttt{[SOS] bird pics [SEP] northern male cardinal [EOS]}}.
    We use \texttt{wordsegment}~\cite{wordsegment} to break subreddit names to words (e.g. itookapicture $\rightarrow$ i took a picture).
    \item \textbf{Training details:}
    We use AdamW~\cite{kingma2015adam,loshchilov2019decoupled} with weight decay $10^{-2}$ and max learning rate $5\times10^{-4}$
    with linear warmup for the first 10K iterations, followed by cosine decay~\cite{loshchilov2016sgdr} to zero.
    We also use label smoothing ($\epsilon_{ls}$ = 0.1)~\cite{szegedy2016rethinking} which has improved language generation for machine translation~\cite{vaswani2017attention}.
    We train for 1.5M iterations with total batch size 256 across 8$\times$ 2080Ti GPUs.
\end{compactitem}

We save checkpoints every 2000 iterations, and average the last five checkpoints to use for downstream tasks and image captioning.
All other details remain unchanged from \cite{desai2020virtex}. We have open-sourced all the training code and pre-trained checkpoints, available at \url{https://redcaps.xyz}.

\vspace{-6pt}
\subsection{Transfer learning on downstream vision tasks}
\label{subsec:downstream}
\vspace{-6pt}

We evaluate the quality of visual representations learned from \sbu{}, \gcc{}, and \dsetname{}
by training \virtexii{} models on each, then transferring the visual backbone to image classification and instance segmentation on \textbf{eleven} different downstream datasets.
Our evaluation setup closely follows recent works on self-supervised learning~\cite{he2019moco,chen2020simclr,caron2020swav} and language-supervised~\cite{desai2020virtex,radford2021clip} learning.
We describe the main evaluation settings here; see Appendix \ref{appendix:experiments} for more details.

\begin{table*}[t]
    \centering
    \small
    \setlength\tabcolsep{2pt}
    \parbox[t]{2mm}{\rotatebox[origin=c]{90}{\textbf{Lin. Probe \hspace{3pt} Zero Shot \hspace{32pt}}}}
    \hfill
    \begin{tabularx}{0.98\linewidth}{l cZ cZ cZ cZ cZ cZ cZ cZ cZ}
    \toprule
    \multicolumn{1}{l}{\bf \multirow[b]{2}{*}{\shortstack{Pre-train\\Dataset}}}
    && Pets && Food && Flowers && Cars && Country && SUN && Birdsnap
    && \multicolumn{1}{l}{\bf \multirow[b]{2}{*}{\shortstack{Average\\Accuracy}}} \\
    \cmidrule{3-3} \cmidrule{5-5} \cmidrule{7-7} \cmidrule{9-9}
    \cmidrule{11-11} \cmidrule{13-13} \cmidrule{15-15}
    && \emph{N = 37}  && \emph{N = 101} && \emph{N = 102} && \emph{N = 196}
    && \emph{N = 211} && \emph{N = 397} && \emph{N = 500} \\
    \midrule
    \sbu{}
    && 8.7 && 3.0 && 13.7 && 0.6
    && 0.6 && 14.7 && 1.3 && 7.0\\
    \gcc{}
    && 15.5 && 10.9 && 10.1 && 0.5
    && 0.5 && \textbf{33.3} && 1.6 && 12.0 \\
    \band \dsetname-20
    && 41.8 && \textbf{54.6} && \textbf{33.5} && \textbf{3.2}
    && 2.3 && 23.9 && \textbf{11.8} && \textbf{28.1} \\
    \band \dsetname{}
    && \textbf{42.4} && 53.8 && 26.2 && 3.1
    && \textbf{3.6} && 26.8 && 8.3 && 26.8 \\
    \midrule
    \sbu{}
    && 61.8 && 48.5 && 80.3 && 22.2
    && 12.0 && 61.3 && 18.6 && 44.5 \\
    \gcc{}
    && 69.9 && 57.3 && 76.6 && 25.2
    && 12.8 && \textbf{70.0} && 16.1 && 46.8 \\
    \band \dsetname-20
    && \textbf{87.0} && 79.1 && 85.9 && 39.1
    && 11.6 && 63.6 && \textbf{30.6} && 55.2 \\
    \band \dsetname{}
    && 85.0 && \textbf{80.8} && \textbf{86.3} && \textbf{43.9}
    && \textbf{13.6} && 67.3 && 28.1 && \textbf{56.7} \\
    \bottomrule
  \end{tabularx}
  \caption{
      \textbf{Transfer learning: zero-shot and linear probe.}
      We train \virtexii{} models on different image-text datasets,
      then transfer the learned features to seven downstream classification datasets (\emph{N = \#classes}).
      Models trained on \dsetname{} perform best on all datasets except one.
  }
  \label{tab:transfer1}
  \vspace{-16pt}
\end{table*}

\paragraph{Zero-shot image classification:}
Training with language supervision enables \emph{zero-shot} transfer to downstream tasks without \emph{any} task-specific training~\cite{radford2021clip,li2017ngrams}.
We evaluate the utility of different datasets for representation learning by comparing zero-shot performance on seven classification datasets:
Oxford-IIIT Pets~\cite{parkhi2012pets}, Food-101~\cite{bossard2014food}, Flowers-102~\cite{nilsback2008flowers}, Stanford Cars~\cite{krause2013cars}, Country-211~\cite{radford2021clip}, and SUN-397~\cite{xiao2010sun397}, and Birdsnap~\cite{berg2014birdsnap}.
Inspired by CLIP~\cite{radford2021clip}, we perform zero-shot classification by designing one \emph{prompt} per category in the target dataset and ranking the log-probabilities predicted by the trained captioning model for each prompt, averaging predictions from the forward and backward Transformers.
For \sbu{} and \gcc{} we follow CLIP and use the prompt \inlinecap{\small [SOS] a photo of a/an \_ [EOS]};
for \dsetname{} we adjust to the training setup and use a prompt with prefixed subreddit --
\inlinecap{\small [SOS] i took a picture [SEP] itap of a/an \_ [EOS]}.

Results are shown in \Cref{tab:transfer1} \textcolor{Red}{(top)}.
\virtexii{} models trained on \dsetname{} outperform those trained on \sbu{} and \gcc{} by \emph{wide} margins on \textbf{six} out of seven datasets.
This not due to \dsetname{}'s larger size: models trained on \dsetname{}-20 also outperform those trained on \gcc{}.

\paragraph{Linear probe image classification:}
We also evaluate image classification on these datasets by training linear models over \emph{frozen} visual features.
Our evaluation details exactly follow CLIP -- we use \texttt{scikit-learn}~\cite{scikit-learn} logistic regression with L-BFGS. We train for 1K iterations, and search L2 regularization $\lambda$ over 96 logarithmic spaced values in $[10^{-6}, 10^6]$ by validating on held-out 10\% training data.
Results are shown in \Cref{tab:transfer1} \textcolor{Red}{(bottom)} with similar trends as zero-shot transfer.

\paragraph{Comparison with CLIP:}
Despite improvements over \sbu{} and \gcc{}, our absolute zero-shot performance falls behind CLIP
(e.g Food-101 top-1 with ResNet-50 -- 81.1 vs. 54.6).
Their results are not comparable, as CLIP uses a different architecture (contrastive vs autoregressive), deeper transformer (12 vs 6 layers), larger dataset (400M vs 12M instances), longer training (12.8B image updates vs 384M), and prompt ensembling.
Our goal is not to achieve state-of-the-art performance, but instead to compare impact of different data sources on the quality of learned visual features.

\begin{wraptable}{r}{0.52\linewidth}
    \vspace{-10pt}
    \small
    \setlength\tabcolsep{1pt}

    \begin{tabularx}{\linewidth}{l c ZZZ c Z c Z c Z}
    \toprule
    \multicolumn{1}{l}{\bf \multirow[b]{2}{*}{\shortstack{Pre-train\\Dataset}}}
    && \multicolumn{3}{c}{ImageNet Top-1} && VOC && COCO && LVIS \\
    \cmidrule{3-5} \cmidrule{7-7} \cmidrule{9-9} \cmidrule{11-11}
    && \shortstack{Zero\\shot} & \shortstack{Linear\\Cls.} & \shortstack{k-NN\\(k=20)}
    && \shortstack{Cls.\\mAP} && \shortstack{Segm.\\AP} && \shortstack{Segm.\\AP} \\
    \midrule
    \sbu{}            &&  5.2 & 45.5 & 38.7 && 85.0 && 36.5 && 22.0 \\
    \gcc{}            && 20.7 & \textbf{53.9} & 45.4 && 87.0 && \textbf{37.2} && 22.9 \\
    \band \dsetname{} && \textbf{22.7} & 53.4 & \textbf{52.0} && \textbf{87.5} && 37.0 && \textbf{23.0} \\
    \bottomrule
    \end{tabularx}
    \vspace{-5pt}
    \caption{
        \textbf{Additional tasks:}
        \dsetname{} trained model matches or exceeds models trained on \sbu{}/\gcc{}.
    }
    \label{tab:transfer2}
\end{wraptable}

\paragraph{Other tasks:}
We evaluate on standard transfer tasks with four other datasets: \voc{} and \imagenet{} linear classification with \emph{frozen} features
and instance segmentation~\cite{he2017mask} on COCO~\cite{lin2014microsoft} and LVIS~\cite{gupta2019lvis} with \emph{end-to-end fine-tuning} of Mask R-CNN.
These tasks follow the same setup as \cite{desai2020virtex}.
On ImageNet, we also perform  $k$ nearest neighbor classification ($k{=}20$), following \cite{wu2018npid,caron2021dino}, and zero-shot classification as described above.
Results are shown in \Cref{tab:transfer2}.
All models perform similarly on fine-tuning tasks (COCO and LVIS), while \dsetname{} trained model gains on tasks involving minimal or no fine-tuning -- k-NN (52.0 vs 45.4) and zero-shot (22.7 vs 20.7) on ImageNet, and linear classification on VOC (87.5 vs 87.0).

\subsection{Image captioning}
\label{subsec:captioning}

We hope that the human interaction flavored data of \dsetname{} enables more human-like and \emph{conversational} image captioning models.
We use \virtexii{} pre-trained models for image captioning -- we use nucleus sampling~\cite{holtzman2020nucleus} with nucleus size 0.9 to decode a caption from the forward Transformer.
In this section, we demonstrate all results on an additional \emph{held-out test set} of 1K instances sampled randomly from image posts
submitted to our selected subreddits in the first week of 2021.

\begin{wrapfigure}{r}{0.23\linewidth}
    \vspace{-10pt}
    \footnotesize
    \begin{tabularx}{\linewidth}{XX}
        \shortstack{\dsetname{}\\vs.} & \shortstack{\dsetname{}\\preferred} \\
        \midrule
        \gcc{} & 63.3\% \\
        Human & 41.6\% \\
    \end{tabularx}
    \vspace{-10pt}
\end{wrapfigure}

\paragraph{Evaluating caption predictions:}
Automatic captioning evaluation metrics correlate poorly with human judgement~\cite{vedantam2015cider,anderson2016spice}.
We thus evalute caption predictions via user studies.
We sample captions from models trained on \dsetname{} and \gcc{}, then present crowd workers with the image and both captions.
Workers are told that one caption is written by a human and the other machine-generated, and asked to guess which is human-written.
We take a majority vote among three workers for each of our 1K test images.
Results are shown to the right -- workers preferred captions from the \dsetname{}-trained model for 633/1000 images.
We run a similar study to compare against ground-truth captions, and workers still prefer generated captions for 416/1000 images.
Some qualitative results are shown in \Cref{fig:human_eval}; more are shown in Appendix (\Cref{fig:supp_human_eval}).

\begin{figure}[t]
    \vspace{-10pt}
    \footnotesize
    \setlength \tabcolsep{1pt}
    \renewcommand{\arraystretch}{1.2}

    \newcommand{\blueband}{\rowcolor{Blue!10}}
    \newcommand{\greenband}{\rowcolor{ForestGreen!10}}

    \newcolumntype{Y}{>{\centering\arraybackslash}X}
    \begin{tabularx}{\linewidth}{lYYYYc}
        ~ &
        \includegraphics[width=0.223\textwidth, height=70pt]{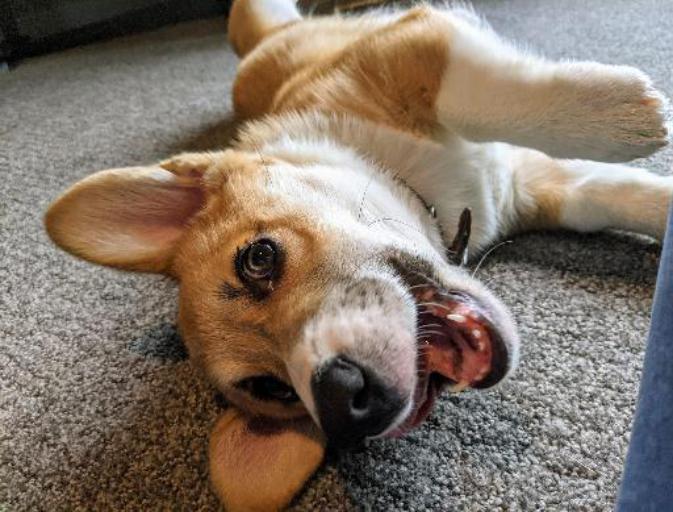} &
        \includegraphics[width=0.223\textwidth, height=70pt]{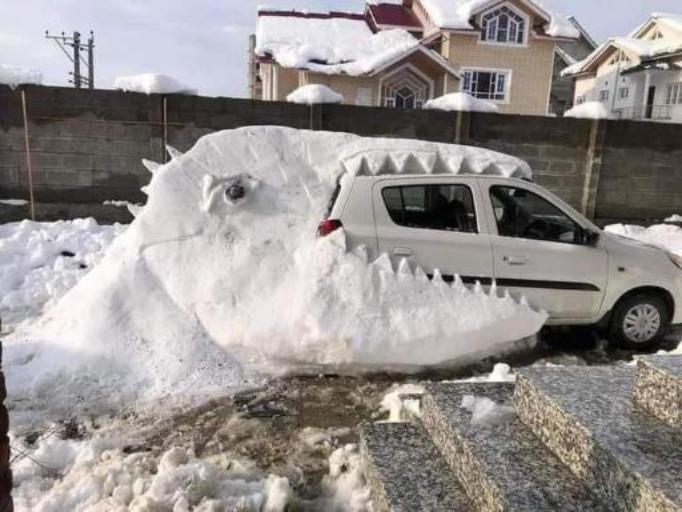} &
        \includegraphics[width=0.223\textwidth, height=70pt]{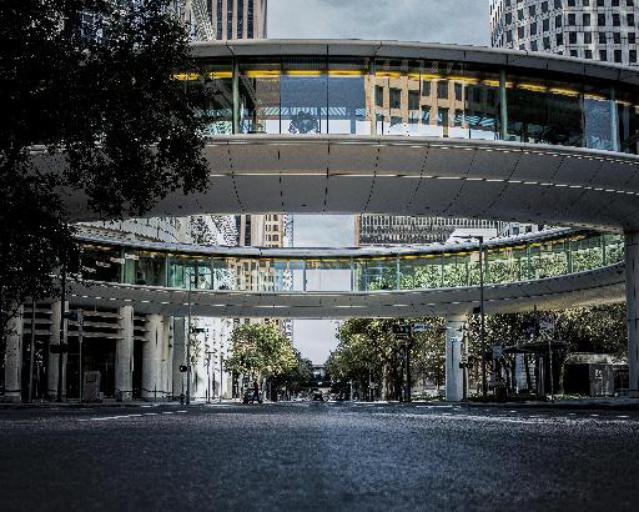} &
        \includegraphics[width=0.223\textwidth, height=70pt]{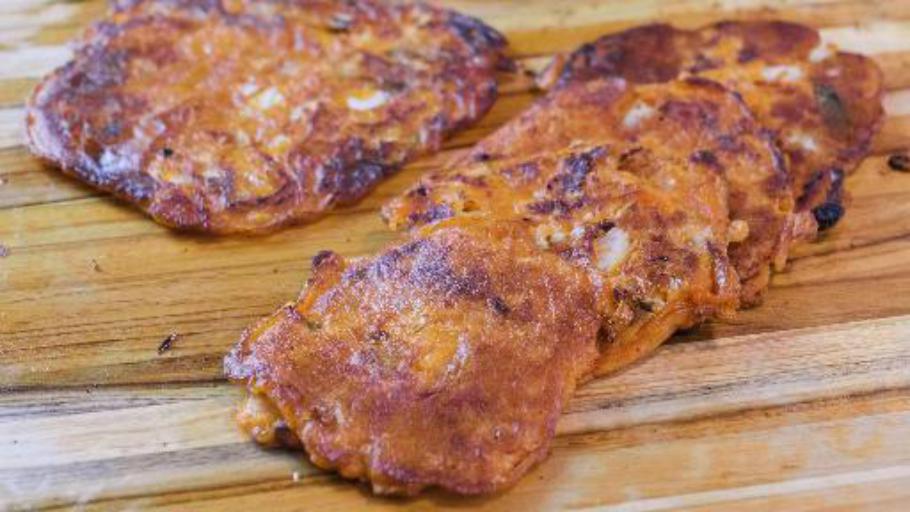} &
        \\
        \blueband \gcc{} &
        animal lying on the ground &
        a car is completely covered in snow. &
        the building is a-story polished concrete floor. &
        \ul{how to cook a rack of ribs} &
        \\
        \greenband \dsetname{} &
        \ttbf{r/lookatmydog:} \ul{my little guy} &
        \ttbf{r/mildlyinteresting:} \ul{this snow sculpture} &
        \ttbf{r/pics:} \ul{a building in singapore} &
        \ttbf{r/foodporn:} homemade pizza &
        \\
    \end{tabularx}
    \caption{
        \textbf{Human evaluation: \gcc{} vs. \dsetname{}.}
        We decode image captions from \virtexii{} models trained on \gcc{} and \dsetname{}.
        We show both captions (excluding subreddit names) to three crowd workers and ask them to guess which is more likely to be written by a human.
        All three workers chose the underlined caption for each of the displayed images.
        We found that workers preferred organic references (\textcolor{ForestGreen}{little guy} vs \textcolor{RoyalBlue}{animal}), witty remarks (\textcolor{ForestGreen}{snow sculpture}), and specific mentions (\textcolor{ForestGreen}{singapore}) by the \dsetname{}-trained model.
        Among negative cases are mostly instances where \dsetname{}-trained models make blatant errors in identifying common visual objects (e.g. \textcolor{ForestGreen}{pizza}).
        }
    \vspace{-12pt}
    \label{fig:human_eval}
\end{figure}

\begin{figure}[t]
    \vspace{-5pt}
    \footnotesize
    \setlength \tabcolsep{2pt}
    \renewcommand{\arraystretch}{1.2}

    \newcommand{\blueband}{\rowcolor{Blue!10}}
    \newcommand{\brownband}{\rowcolor{Brown!10}}
    \newcommand{\greenband}{\rowcolor{ForestGreen!10}}

    \newcolumntype{Y}{>{\centering\arraybackslash}X}
    \begin{tabularx}{\linewidth}{YYYYc}
        \includegraphics[width=0.243\textwidth, height=70pt]{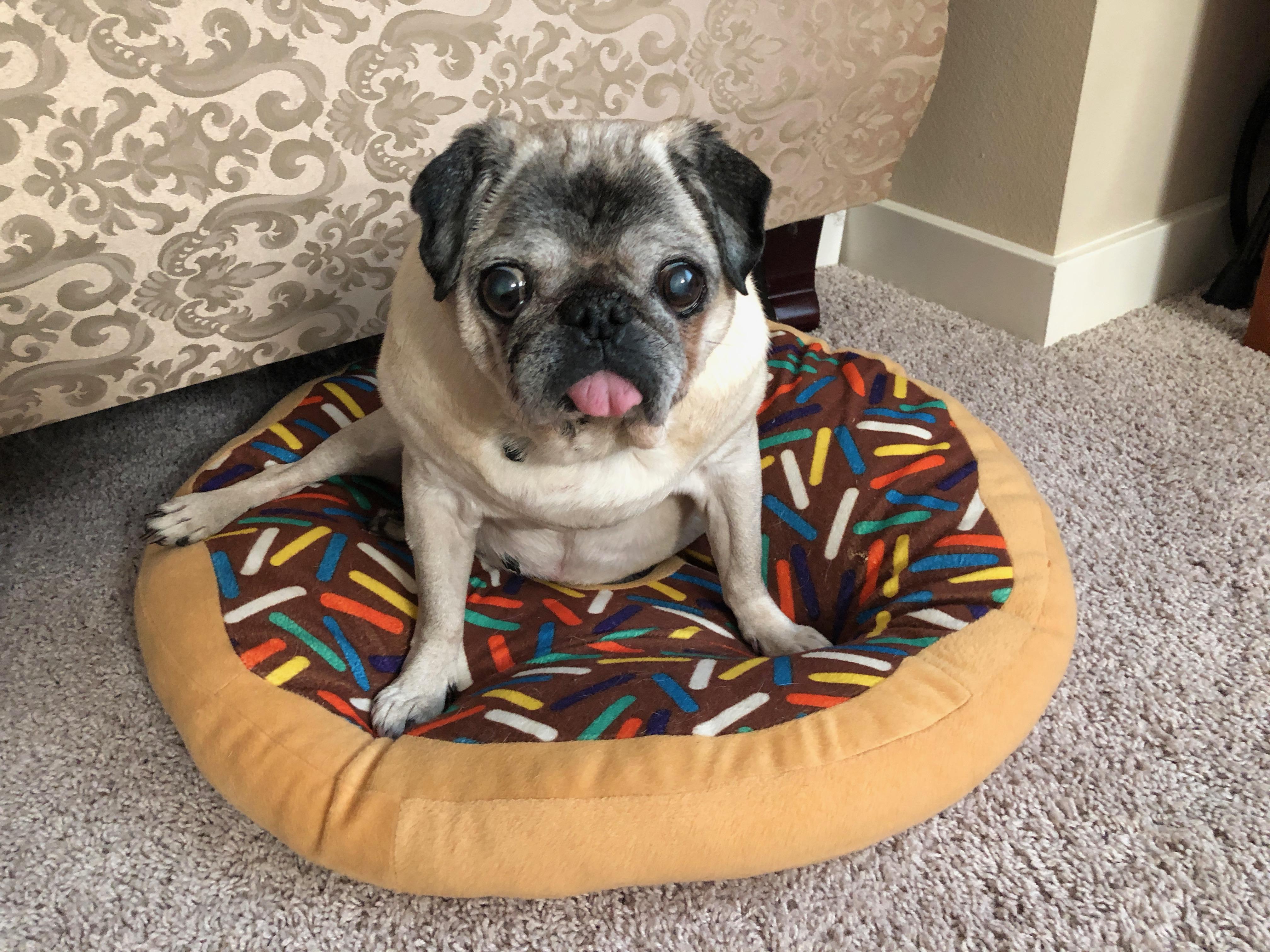} &
        \includegraphics[width=0.243\textwidth, height=70pt]{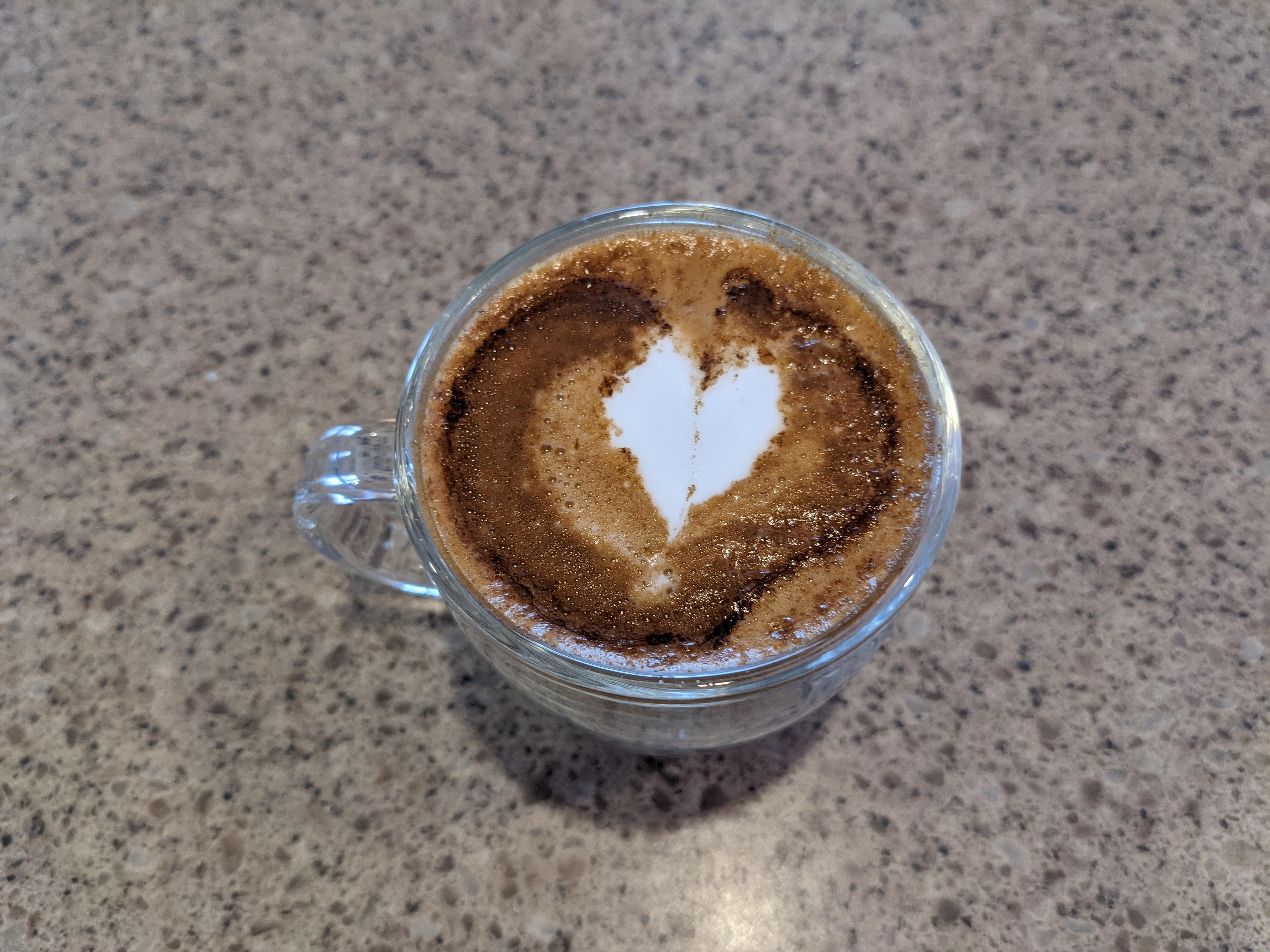} &
        \includegraphics[width=0.243\textwidth, height=70pt]{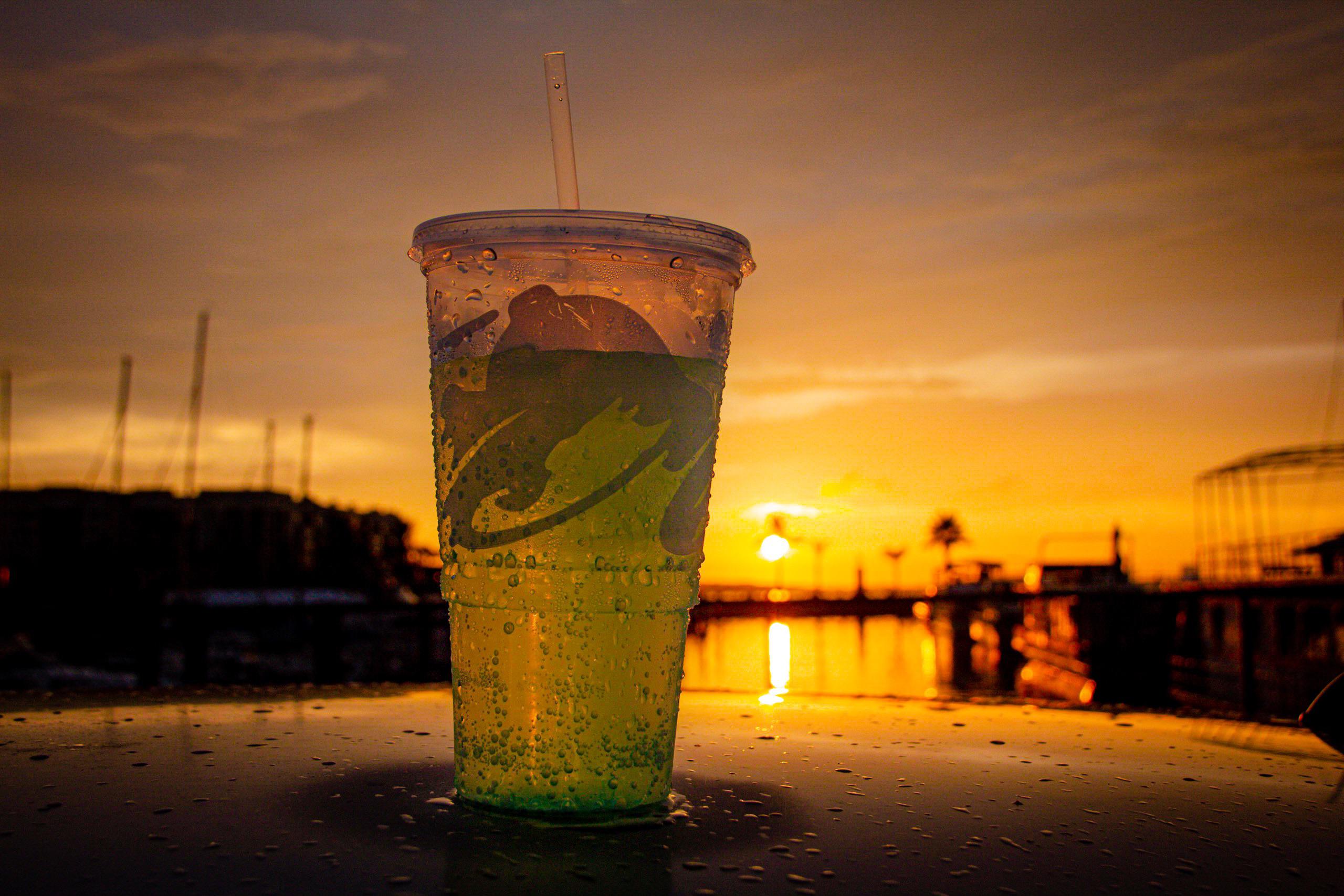} &
        \includegraphics[width=0.243\textwidth, height=70pt]{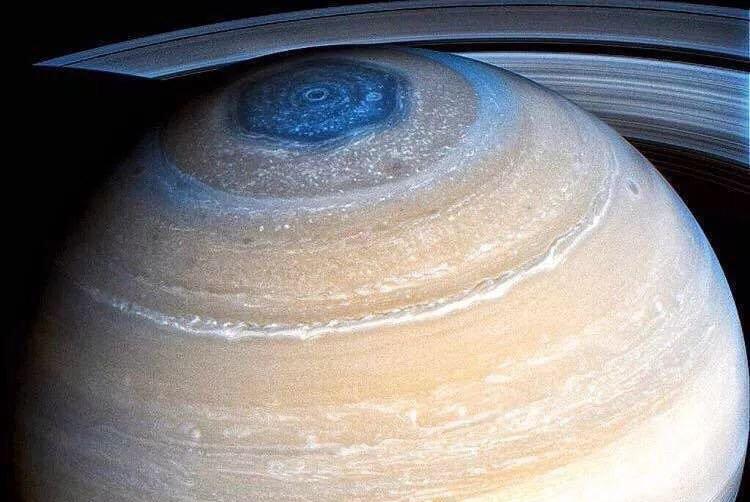} &
        \\
        \blueband \ttbf{r/itookapicture:} itap of my dog. &
        \ttbf{r/itookapicture:} itap of my coffee &
        \ttbf{r/earthporn:} sunset in venice, italy &
        \ttbf{r/earthporn:} saturn's north pole. &
        \\
        \brownband \ttbf{r/absoluteunits:} this absolute unit of a pug &
        \ttbf{r/absoluteunits:} this absolute unit of a coffee. &
        \ttbf{r/food:} a cold beer on the beach. &
        \ttbf{r/food:} the clearest image of saturn &
        \\
        \greenband \ttbf{r/somethingimade:} i made a bed for my pug. &
        \ttbf{r/somethingimade:} i made a heart latte. &
        \ttbf{r/pics:} shot from a beach house! &
        \ttbf{r/pics:} the clearest image of saturn ever taken &
        \\
    \end{tabularx}
    \caption{
        \textbf{Subreddit-controlled caption style.}
        We prompt the \virtexii{} model trained on \dsetname{} with subreddit names while decoding captions.
        We observe that such conditioning captures subtle linguistic structures
        (\ttbf{r/itookapicture}: \textcolor{RoyalBlue}{itap of ...}, \ttbf{r/somethingimade}: \textcolor{ForestGreen}{i made...}).
        or changes the main subject of caption (\ttbf{r/earthporn}: \textcolor{RoyalBlue}{venice}, \ttbf{r/food}: \textcolor{Brown}{cold beer}).
        However, for completely unrelated images (saturn), the model tends to ignore the conditioning while generating captions.
    }
    \vspace{-5pt}
    \label{fig:style_captioning}
\end{figure}

\paragraph{Subreddit-conditioned generation:}
Captions from different subreddits have distinct styles,
focusing on different image aspects or using community-specific jargon.
We use this observation to generate captions with distinct styles
by prompting a \dsetname{}-trained model with \emph{different} subreddits.
Figure~\ref{fig:style_captioning} shows examples of such diverse captions for images;
see Appendix (\Cref{fig:caption_style_supp1}) for more.

\section{Related work}
\label{sec:related}

\dsetname{} is directly related to many recent efforts on building large datasets of image-text pairs from the internet without expensive human annotation.
Two notable datasets are~\sbu{}~\cite{ordonez2011im2text} and Conceptual Captions~\cite{sharma2018conceptual}.
Originally intended for image-text retrieval and image captioning,
they are now widely used for training generic \vnl{} representations~\cite{tan2019lxmert,lu2019vilbert,li2019visualbert,su2019vl,li2020unicoder,chen2019uniter,zhou2020vlp,li2020oscar,huang2020pixelbert,kim2021vilt}
that transfer to downstream tasks like visual question answering~\cite{antol2015vqa,zhu2016visual7w,hudson2019gqa}, referring expressions~\cite{kazemzadeh2014referitgame}, and visual reasoning~\cite{suhr2019corpus,zellers2019recognition}.
More recent works build larger datasets specifically for \vnl{} pre-training, e.g. LAIT~\cite{qi2020imagebert}, Conceptual-12M~\cite{changpinyo2021conceptual}, and Wikipedia-ImageText~\cite{srinivasan2021wit}.
Similar to these datasets, \dsetname{} offers rich semantic data for pre-training applications.
However, our choice of data source and hence the data quality is unique.

Image-text datasets are now also used for learning visual features.
\citet{li2017ngrams} trained visual N-gram models on YFCC-100M~\cite{yfcc100m};
\cite{desai2020virtex,bulent2020icmlm} learn features from COCO Captions~\cite{chen2015microsoft} that are competitive with supervised ImageNet training~\cite{krizhevsky2012imagenet,he2016deep} on many downstream tasks~\cite{everingham2009voc,russakovsky2015imagenet,lin2014microsoft,gupta2019lvis,van2018inaturalist}, and \cite{radford2021clip,jia2021scaling} scale up to very larger non-public datasets that are larger than \dsetname{}.

A core motivation for collecting image-text data is scaling to larger datasets without bearing annotation costs.
Related to this goal are efforts that learn from large quantities of noisy non-text labels for web images such as WebVision~\cite{li2017webvision}, YFCC-100M~\cite{yfcc100m}, JFT-300M~\cite{hinton2015distilling,chollet2017xception}, and Instagram-3.5B~\cite{mahajan2018exploring}.

\section{Conclusion}
\label{sec:conclusion}

This paper has introduced \dsetname{}, a large-scale dataset of images and captions collected from Reddit.
As a source of data, Reddit is appealing:
text and image are both created and shared by people, for the explicit purpose of starting a discussion with other people, leading to natural and varied content.
Its subreddit structure allows manually curation of our dataset's content without labeling individual instances.
We utilize this structure to collect a dataset focused on animals, objects, scenery, and activities, and specifically aim to minimize the appearance of people.
We have shown that \dsetname{} is useful for learning visual representations that transfer to many downstream tasks, including zero-shot settings that use no task-specific training data.
We have also shown that \dsetname{} can be used to learn image captioning models that generate high-quality text of multiple styles.

\dsetname{} is not without flaws.
We have tried to minimize problematic content through subreddit curation and automated filtering, but the unfathomable nature of large data means that \dsetname{} may contain a small number of instances with NSFW images or harmful language.
Reddit's demographic biases mean that \dsetname{} may not equally represent all groups.
Users should carefully consider these limitations for any new tasks developed on \dsetname{},
and should be especially wary of applications that make predictions about people.
Despite these limitations, we hope that \dsetname{} will help enable a wide variety of new applications and advances in vision and language.

\clearpage

\section*{Acknowledgments}

We thank Mohit Virli for suggestions on the project website.
We thank Mohamed El Banani, Nilesh Kulkarni, Stefan Lee, Ramprasaath Selvaraju, Ramakrishna Vedantam, and Erik Wijmans for helpful discussions and feedback on the paper.
We thank Priya Goyal and Ishan Misra for help related to VISSL codebase.
We thank all anonymous reviewers for constructive feedback during the review phase.
We also thank the UMich ARC-TS team for support with GPU cluster management.

\begin{appendices}
  \section{List of all subreddits in \dsetname{}}
\label{appendix:subreddits}

We curated \dsetname{} from a manually chosen set of \dsetsubs{} subreddits, as described in \Cref{subsec:pipeline}.
All these subreddits are listed below alphabetically with the number of instances in each subreddit.

\begin{table*}[ht]
    \centering
    \setlength{\tabcolsep}{1pt}
    \small

    \newcommand{\brownband}{\rowcolor{Brown!10}}
    \newcolumntype{Y}{>{\raggedright\arraybackslash}X}
    \begin{tabularx}{\linewidth}{Yc c Yc c Yc}
        \brownband \subreddit{abandoned} & 7.0K  && \subreddit{abandonedporn} & 56.2K  && \subreddit{absoluteunits} & 33.9K \\
        \subreddit{airplants} & 8.6K  && \subreddit{alltheanimals} & 1.3K  && \subreddit{amateurphotography} & 14.0K \\
        \brownband \subreddit{amateurroomporn} & 11.6K  && \subreddit{animalporn} & 15.8K  && \subreddit{antiques} & 17.0K \\
        \subreddit{antkeeping} & 3.0K  && \subreddit{ants} & 1.2K  && \subreddit{aquariums} & 139K \\
        \brownband \subreddit{architectureporn} & 14.1K  && \subreddit{artefactporn} & 9.6K  && \subreddit{astronomy} & 2.1K \\
        \subreddit{astrophotography} & 24.6K  && \subreddit{australiancattledog} & 21.4K  && \subreddit{australianshepherd} & 12.5K \\
        \brownband \subreddit{autumnporn} & 2.7K  && \subreddit{averagebattlestations} & 5.6K  && \subreddit{awwducational} & 5.9K \\
        \subreddit{awwnverts} & 7.6K  && \subreddit{axolotls} & 9.7K  && \subreddit{backpacking} & 8.9K \\
        \brownband \subreddit{backyardchickens} & 17.3K  && \subreddit{baking} & 119K  && \subreddit{ballpython} & 19.2K \\
        \subreddit{barista} & 6.6K  && \subreddit{bassfishing} & 6.5K  && \subreddit{battlestations} & 58.4K \\
        \brownband \subreddit{bbq} & 22.3K  && \subreddit{beagle} & 21.4K  && \subreddit{beardeddragons} & 55.1K \\
        \subreddit{beekeeping} & 1.2K  && \subreddit{beerandpizza} & 1.2K  && \subreddit{beerporn} & 95.7K \\
        \brownband \subreddit{beerwithaview} & 8.9K  && \subreddit{beginnerwoodworking} & 8.7K  && \subreddit{bengalcats} & 7.0K \\
        \subreddit{bento} & 4.8K  && \subreddit{bernesemountaindogs} & 6.4K  && \subreddit{berries} & 805 \\
        \brownband \subreddit{bettafish} & 64.7K  && \subreddit{bicycling} & 80.8K  && \subreddit{bikecommuting} & 9.8K \\
        \subreddit{birding} & 21.1K  && \subreddit{birdphotography} & 1.3K  && \subreddit{birdpics} & 29.1K \\
        \brownband \subreddit{birds} & 1.2K  && \subreddit{birdsofprey} & 2.2K  && \subreddit{blackcats} & 84.1K \\
        \subreddit{blacksmith} & 13.3K  && \subreddit{bladesmith} & 9.1K  && \subreddit{boatporn} & 2.8K \\
        \brownband \subreddit{bonsai} & 18.1K  && \subreddit{bookporn} & 4.5K  && \subreddit{bookshelf} & 6.2K \\
        \subreddit{bordercollie} & 21.9K  && \subreddit{bostonterrier} & 28.2K  && \subreddit{botanicalporn} & 13.4K \\
        \brownband \subreddit{breadit} & 71.6K  && \subreddit{breakfast} & 2.2K  && \subreddit{breakfastfood} & 3.8K \\
        \subreddit{bridgeporn} & 2.4K  && \subreddit{brochet} & 3.2K  && \subreddit{budgetfood} & 1.6K \\
        \brownband \subreddit{budgies} & 1.3K  && \subreddit{bulldogs} & 24.1K  && \subreddit{burgers} & 10.7K \\
        \subreddit{butterflies} & 4.5K  && \subreddit{cabinporn} & 2.7K  && \subreddit{cactus} & 36.5K \\
        \brownband \subreddit{cakedecorating} & 14.0K  && \subreddit{cakewin} & 4.8K  && \subreddit{cameras} & 3.3K \\
        \subreddit{camping} & 21.4K  && \subreddit{campingandhiking} & 25.5K  && \subreddit{carnivorousplants} & 1.3K \\
        \brownband \subreddit{carpentry} & 4.1K  && \subreddit{carporn} & 102K  && \subreddit{cassetteculture} & 12.2K \\
        \subreddit{castiron} & 33.6K  && \subreddit{castles} & 7.0K  && \subreddit{casualknitting} & 3.1K \\
        \brownband \subreddit{catpictures} & 51.9K  && \subreddit{cats} & 643K  && \subreddit{ceramics} & 4.8K \\
        \subreddit{chameleons} & 7.9K  && \subreddit{charcuterie} & 3.0K  && \subreddit{cheese} & 5.0K \\
        \brownband \subreddit{cheesemaking} & 1.7K  && \subreddit{chefit} & 1.6K  && \subreddit{chefknives} & 8.7K \\
        \subreddit{chickens} & 9.6K  && \subreddit{chihuahua} & 36.2K  && \subreddit{chinchilla} & 5.6K \\
        \brownband \subreddit{chinesefood} & 1.8K  && \subreddit{churchporn} & 2.1K  && \subreddit{cider} & 2.4K \\
        \subreddit{cityporn} & 56.9K  && \subreddit{classiccars} & 14.4K  && \subreddit{cockatiel} & 12.1K \\
        \brownband \subreddit{cocktails} & 25.0K  && \subreddit{coffeestations} & 1.5K  && \subreddit{coins} & 45.0K \\
        \subreddit{cookiedecorating} & 3.7K  && \subreddit{corgi} & 64.7K  && \subreddit{cornsnakes} & 3.4K \\
        \brownband \subreddit{cozyplaces} & 44.9K  && \subreddit{crafts} & 44.0K  && \subreddit{crestedgecko} & 5.2K \\
        \subreddit{crochet} & 125K  && \subreddit{crossstitch} & 63.6K  && \subreddit{crows} & 1.1K \\
        \brownband \subreddit{crystals} & 24.0K  && \subreddit{cupcakes} & 2.3K && \subreddit{dachshund} & 47.0K \\
        \subreddit{damnthatsinteresting} & 28.4K  && \subreddit{desertporn} & 1.2K  && \subreddit{designmyroom} & 6.3K \\
        \brownband \subreddit{desksetup} & 1.1K  && \subreddit{dessert} & 3.2K  && \subreddit{dessertporn} & 9.5K \\
        \subreddit{diy} & 19.4K  && \subreddit{dobermanpinscher} & 8.1K  && \subreddit{doggos} & 18.6K \\
        \brownband \subreddit{dogpictures} & 120K  && \subreddit{drunkencookery} & 5.9K  && \subreddit{duck} & 4.7K \\
        \subreddit{dumpsterdiving} & 4.4K  && \subreddit{earthporn} & 262K  && \subreddit{eatsandwiches} & 20.5K \\
        \brownband \subreddit{embroidery} & 38.5K  && \subreddit{entomology} & 6.9K  && \subreddit{equestrian} & 5.2K \\
        \end{tabularx}
\end{table*}
\clearpage

\begin{table*}[h]
    \centering
    \setlength{\tabcolsep}{1pt}
    \small

    \newcommand{\brownband}{\rowcolor{Brown!10}}
    \newcolumntype{Y}{>{\raggedright\arraybackslash}X}
    \begin{tabularx}{\linewidth}{Yc c Yc c Yc}
        \subreddit{espresso} & 8.5K  && \subreddit{exposureporn} & 10.2K  && \subreddit{eyebleach} & 80.9K \\
        \brownband \subreddit{f1porn} & 12.9K  && \subreddit{farming} & 4.7K  && \subreddit{femalelivingspace} & 947 \\
        \subreddit{fermentation} & 10.6K  && \subreddit{ferrets} & 26.2K  && \subreddit{fireporn} & 1.7K \\
        \brownband \subreddit{fish} & 2.9K  && \subreddit{fishing} & 51.0K  && \subreddit{flowers} & 20.8K \\
        \subreddit{flyfishing} & 19.1K  && \subreddit{food} & 393K  && \subreddit{foodporn} & 202K \\
        \brownband \subreddit{foraging} & 9.5K  && \subreddit{fossilporn} & 1.7K  && \subreddit{fountainpens} & 52.8K \\
        \subreddit{foxes} & 7.7K  && \subreddit{frenchbulldogs} & 12.2K  && \subreddit{frogs} & 14.8K \\
        \brownband \subreddit{gardening} & 208K  && \subreddit{gardenwild} & 1.0K  && \subreddit{geckos} & 5.9K \\
        \subreddit{gemstones} & 1.5K  && \subreddit{geologyporn} & 1.9K  && \subreddit{germanshepherds} & 46.0K \\
        \brownband \subreddit{glutenfree} & 2.9K  && \subreddit{gold} & 1.3K  && \subreddit{goldenretrievers} & 42.4K \\
        \subreddit{goldfish} & 3.9K  && \subreddit{greatpyrenees} & 8.8K  && \subreddit{grilledcheese} & 13.4K \\
        \brownband \subreddit{grilling} & 12.6K  && \subreddit{guineapigs} & 56.8K  && \subreddit{gunporn} & 17.5K \\
        \subreddit{guns} & 99.1K  && \subreddit{hamsters} & 26.9K  && \subreddit{handtools} & 3.2K \\
        \brownband \subreddit{healthyfood} & 8.2K  && \subreddit{hedgehog} & 1.7K  && \subreddit{helicopters} & 3.2K \\
        \subreddit{herpetology} & 9.7K  && \subreddit{hiking} & 41.6K  && \subreddit{homestead} & 9.3K \\
        \brownband \subreddit{horses} & 16.3K  && \subreddit{hotpeppers} & 27.8K  && \subreddit{houseplants} & 182K \\
        \subreddit{houseporn} & 2.8K  && \subreddit{husky} & 35.9K  && \subreddit{icecreamery} & 1.1K \\
        \brownband \subreddit{indoorgarden} & 29.0K  && \subreddit{infrastructureporn} & 7.0K  && \subreddit{insects} & 20.4K \\
        \subreddit{instantpot} & 2.8K  && \subreddit{interestingasfuck} & 73.7K  && \subreddit{interiordesign} & 6.7K \\
        \brownband \subreddit{itookapicture} & 327K  && \subreddit{jellyfish} & 713  && \subreddit{jewelry} & 3.5K \\
        \subreddit{kayakfishing} & 4.8K  && \subreddit{kayaking} & 9.9K  && \subreddit{ketorecipes} & 22.3K \\
        \brownband \subreddit{knifeporn} & 2.5K  && \subreddit{knives} & 63.9K  && \subreddit{labrador} & 25.1K \\
        \subreddit{leathercraft} & 16.0K  && \subreddit{leopardgeckos} & 9.0K  && \subreddit{lizards} & 2.4K \\
        \brownband \subreddit{lookatmydog} & 43.2K  && \subreddit{macarons} & 5.3K  && \subreddit{machineporn} & 6.2K \\
        \subreddit{macroporn} & 14.8K  && \subreddit{malelivingspace} & 17.1K  && \subreddit{mead} & 12.4K \\
        \brownband \subreddit{mealprepsunday} & 33.1K  && \subreddit{mechanicalkeyboards} & 156K  && \subreddit{mechanicalpencils} & 5.3K \\
        \subreddit{melts} & 1.2K  && \subreddit{metalworking} & 3.8K  && \subreddit{microgreens} & 1.1K \\
        \brownband \subreddit{microporn} & 1.8K  && \subreddit{mildlyinteresting} & 731K  && \subreddit{mineralporn} & 10.4K \\
        \subreddit{monitors} & 2.2K  && \subreddit{monstera} & 6.9K  && \subreddit{mostbeautiful} & 25.5K \\
        \brownband \subreddit{motorcycleporn} & 6.4K  && \subreddit{muglife} & 4.1K  && \subreddit{mushroomgrowers} & 13.4K \\
        \subreddit{mushroomporn} & 4.7K  && \subreddit{mushrooms} & 5.6K  && \subreddit{mycology} & 83.6K \\
        \brownband \subreddit{natureisfuckinglit} & 61.3K  && \subreddit{natureporn} & 10.1K  && \subreddit{nebelung} & 4.6K \\
        \subreddit{orchids} & 26.4K  && \subreddit{otters} & 2.6K  && \subreddit{outdoors} & 30.2K \\
        \brownband \subreddit{owls} & 3.6K  && \subreddit{parrots} & 38.0K  && \subreddit{pelletgrills} & 4.5K \\
        \subreddit{pens} & 5.0K  && \subreddit{perfectfit} & 19.7K  && \subreddit{permaculture} & 1.3K \\
        \brownband \subreddit{photocritique} & 51.5K  && \subreddit{photographs} & 11.5K  && \subreddit{pics} & 1.9M \\
        \subreddit{pitbulls} & 88.5K  && \subreddit{pizza} & 46.5K  && \subreddit{plantbaseddiet} & 3.7K \\
        \brownband \subreddit{plantedtank} & 44.4K  && \subreddit{plants} & 42.9K  && \subreddit{plantsandpots} & 3.0K \\
        \subreddit{pomeranians} & 7.4K  && \subreddit{pottery} & 9.6K  && \subreddit{pourpainting} & 15.3K \\
        \brownband \subreddit{proplifting} & 17.8K  && \subreddit{pug} & 5.1K  && \subreddit{pugs} & 40.2K \\
        \subreddit{quilting} & 24.1K  && \subreddit{rabbits} & 105K  && \subreddit{ramen} & 10.9K \\
        \brownband \subreddit{rarepuppers} & 150K  && \subreddit{reeftank} & 29.5K  && \subreddit{reptiles} & 33.1K \\
        \subreddit{resincasting} & 3.7K  && \subreddit{roomporn} & 13.9K  && \subreddit{roses} & 3.2K \\
        \brownband \subreddit{rottweiler} & 11.5K  && \subreddit{ruralporn} & 9.0K  && \subreddit{sailing} & 10.5K \\
        \subreddit{salsasnobs} & 2.9K  && \subreddit{samoyeds} & 6.8K  && \subreddit{savagegarden} & 14.9K \\
        \brownband \subreddit{scotch} & 32.1K  && \subreddit{seaporn} & 2.2K  && \subreddit{seriouseats} & 8.8K \\
        \subreddit{sewing} & 29.7K  && \subreddit{sharks} & 3.0K  && \subreddit{shiba} & 27.8K \\
        \brownband \subreddit{shihtzu} & 8.9K  && \subreddit{shrimptank} & 14.7K  && \subreddit{siamesecats} & 9.6K \\
        \subreddit{siberiancats} & 2.7K  && \subreddit{silverbugs} & 26.1K  && \subreddit{skyporn} & 36.1K \\
        \brownband \subreddit{sloths} & 5.9K  && \subreddit{smoking} & 38.3K  && \subreddit{snails} & 6.9K \\
        \subreddit{snakes} & 45.4K  && \subreddit{sneakers} & 314K  && \subreddit{sneks} & 17.4K \\
        \brownband \subreddit{somethingimade} & 50.4K  && \subreddit{soup} & 1.5K  && \subreddit{sourdough} & 32.2K \\
        \subreddit{sousvide} & 13.6K  && \subreddit{spaceporn} & 16.3K  && \subreddit{spicy} & 12.4K \\
        \brownband \subreddit{spiderbro} & 16.1K  && \subreddit{spiders} & 41.9K  && \subreddit{squirrels} & 8.1K \\
        \subreddit{steak} & 19.8K  && \subreddit{streetphotography} & 10.1K  && \subreddit{succulents} & 201K \\
        \brownband \subreddit{superbowl} & 7.5K  && \subreddit{supermodelcats} & 33.6K  && \subreddit{sushi} & 13.4K \\
        \subreddit{tacos} & 2.7K  && \subreddit{tarantulas} & 15.0K  && \subreddit{tastyfood} & 2.3K \\
        \brownband \subreddit{tea} & 20.5K  && \subreddit{teaporn} & 1.2K  && \subreddit{tequila} & 2.9K \\
        \subreddit{terrariums} & 7.3K  && \subreddit{thedepthsbelow} & 2.5K  && \subreddit{thriftstorehauls} & 91.4K \\
        \brownband \subreddit{tinyanimalsonfingers} & 3.1K  && \subreddit{tonightsdinner} & 25.7K  && \subreddit{toolporn} & 2.1K \\
        \subreddit{tools} & 21.7K  && \subreddit{torties} & 11.0K  && \subreddit{tortoise} & 5.6K \\
        \brownband \subreddit{tractors} & 2.3K  && \subreddit{trailrunning} & 7.3K  && \subreddit{trains} & 14.2K \\
        \subreddit{trucks} & 30.4K  && \subreddit{turtle} & 9.1K  && \subreddit{underwaterphotography} & 1.2K \\
        \brownband \subreddit{upcycling} & 1.9K  && \subreddit{urbanexploration} & 18.8K  && \subreddit{urbanhell} & 8.1K \\
    \end{tabularx}
\end{table*}
\clearpage

\begin{table}[h]
    \centering
    \setlength{\tabcolsep}{1pt}
    \small

    \newcommand{\brownband}{\rowcolor{Brown!10}}
    \newcolumntype{Y}{>{\raggedright\arraybackslash}X}
    \begin{tabularx}{\linewidth}{Yc c Yc c Yc}
        \subreddit{veganfoodporn} & 18.7K  && \subreddit{veganrecipes} & 9.9K  && \subreddit{vegetablegardening} & 12.1K \\
        \brownband \subreddit{vegetarian} & 9.8K  && \subreddit{villageporn} & 6.4K  && \subreddit{vintage} & 4.4K \\
        \subreddit{vintageaudio} & 12.7K  && \subreddit{vinyl} & 41.7K  && \subreddit{volumeeating} & 2.1K \\
        \brownband \subreddit{watches} & 64.2K  && \subreddit{waterporn} & 9.6K  && \subreddit{weatherporn} & 1.8K \\
        \subreddit{wewantplates} & 17.0K  && \subreddit{wildernessbackpacking} & 3.1K  && \subreddit{wildlifephotography} & 16.3K \\
        \brownband \subreddit{wine} & 12.7K  && \subreddit{winterporn} & 7.0K  && \subreddit{woodcarving} & 6.3K \\
        \subreddit{woodworking} & 112K  && \subreddit{workbenches} & 2.8K  && \subreddit{workspaces} & 1.5K \\
        \brownband \subreddit{yarnaddicts} & 2.6K  && \subreddit{zerowaste} & 7.7K &&& \\
    \end{tabularx}
\end{table}

\section{User studies interface for caption evaluation}

In \Cref{subsec:captioning}, we conducted user studies to evaluate the quality of caption predictions from \virtexii{} models trained on \gcc{} and \dsetname{}.
Here are some additional details of the evaluation procedure.
We conduct the user study on the Amazon Mechanical Turk (AMT).
The task is framed as a guessing game -- we mention the crowd-workers that an AI bot is trying to impersonate humans by generating its own image captions.
We set the price of this task as \$0.3 for a batch of 5 images and obtain worker choices for 1K images, 3 workers per image.
Refer the detailed instructions in \Cref{fig:amt_interface} below.
Our final accuracy from this evaluation shows that humans preferred \dsetname{} pre-trained model over \gcc{} for 633/1000 images.

\begin{figure}[h]
  \centering
  \includegraphics[width=\linewidth]{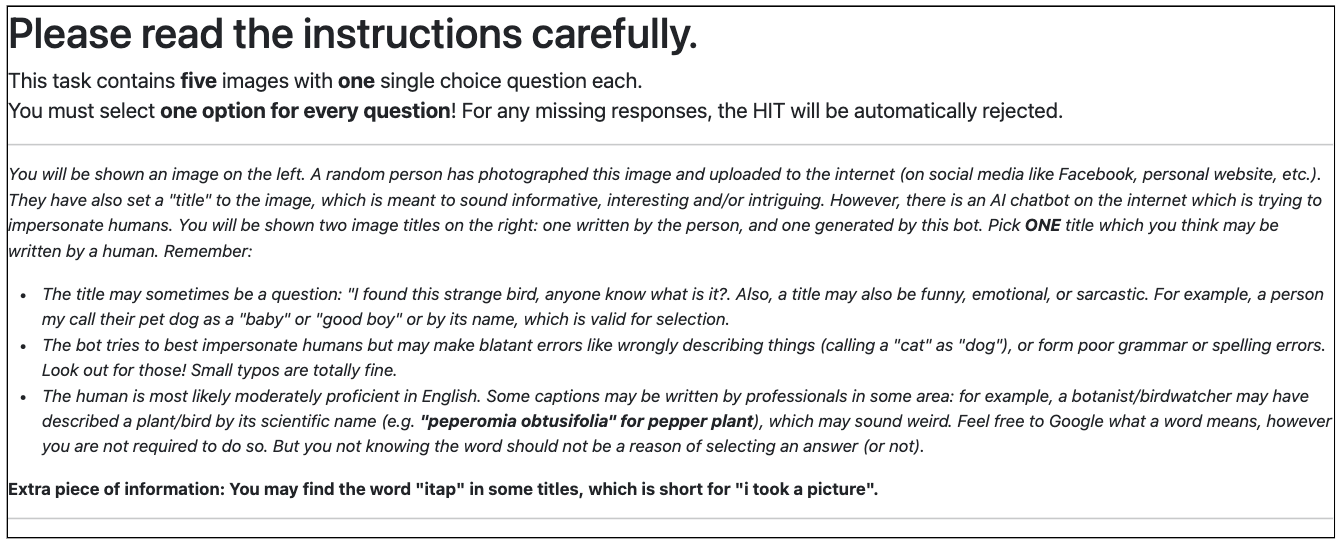}
  \includegraphics[width=\linewidth]{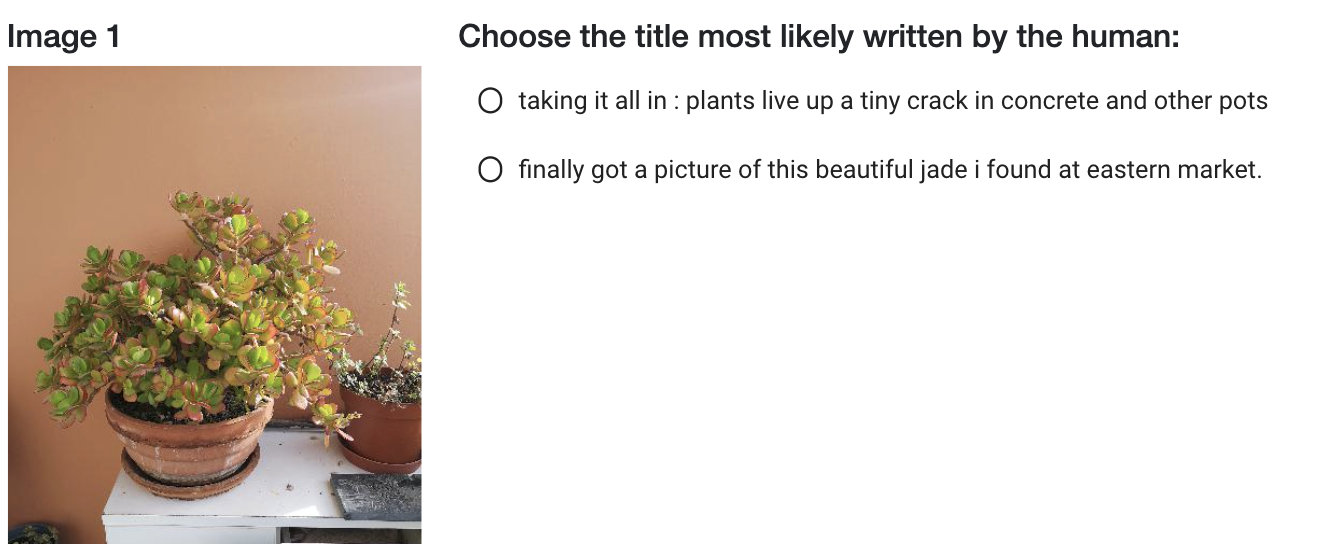}
  \caption{
    \textbf{User studies interface:} Amazon Mechanical Turk task interface -- \emph{instructions and example question} -- for user studies aimed at evaluating the quality of caption predictions from \virtexii{} models trained on \gcc{} vs \dsetname{}.
  }
  \label{fig:amt_interface}
\end{figure}
\clearpage

\section{Qualitative examples: \gcc{} vs \dsetname{}}
\begin{figure}[h]
    \footnotesize
    \setlength \tabcolsep{3pt}
    \renewcommand{\arraystretch}{1.5}

    \newcommand{\blueband}{\rowcolor{Blue!10}}
    \newcommand{\greenband}{\rowcolor{ForestGreen!10}}

    \newcolumntype{Y}{>{\centering\arraybackslash}X}
    \begin{tabularx}{\linewidth}{lYYYYc}
        ~ &
        \includegraphics[width=0.22\textwidth]{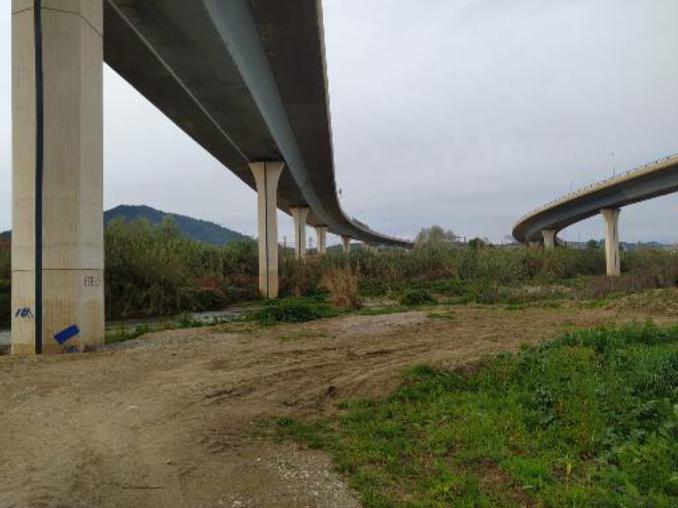} &
        \includegraphics[width=0.22\textwidth]{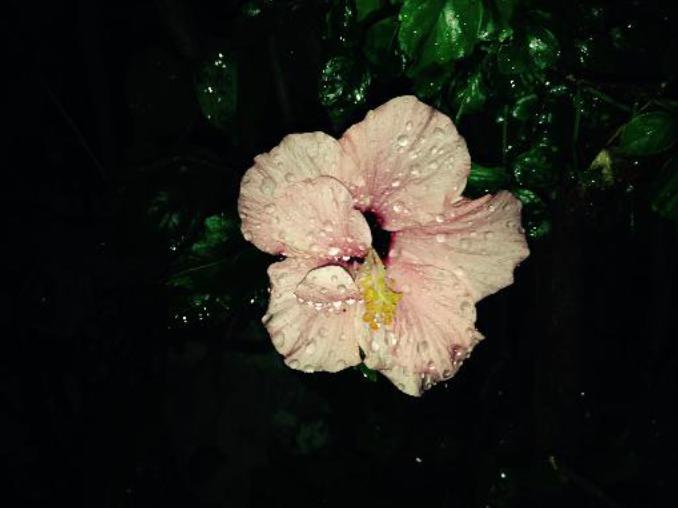} &
        \includegraphics[width=0.22\textwidth]{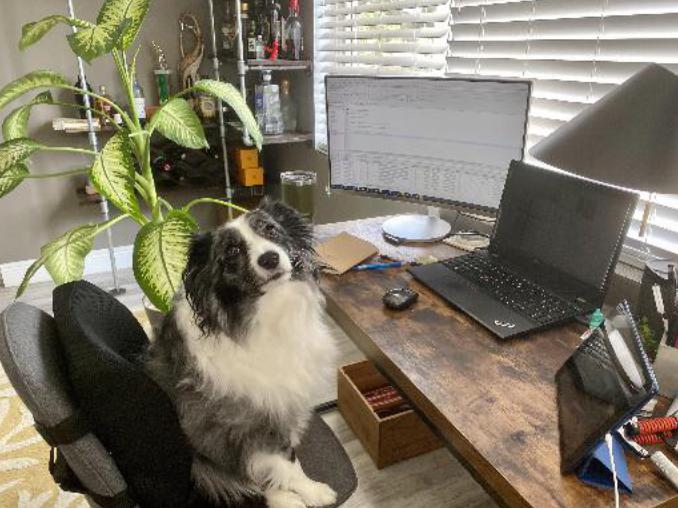} &
        \includegraphics[width=0.22\textwidth]{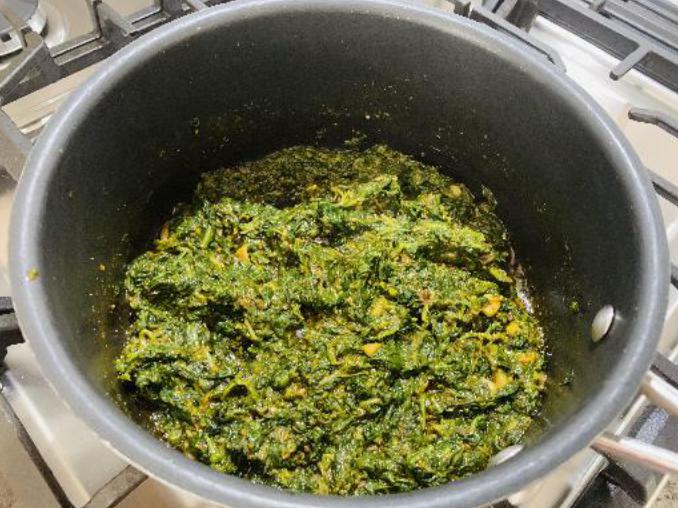} &
        \\
        \blueband \gcc{} &
        \ul{the road leading to the mountains} &
        hibiscus flower in the dark &
        person , the dog , at the office &
        biological variety uncertain future produce slalom &
        \\
        \greenband \dsetname{} &
        \ttbf{r/mildlyinteresting:} this bridge in japan &
        \ttbf{r/pics:} \ul{i took this picture of a hibiscus flower at night} &
        \ttbf{r/mechanicalkeyboards:} \ul{my dog is helping me work from home} &
        \ttbf{r/tea:} \ul{my first time making matcha green tea!} &
        \\
        \\
        ~ &
        \includegraphics[width=0.22\textwidth]{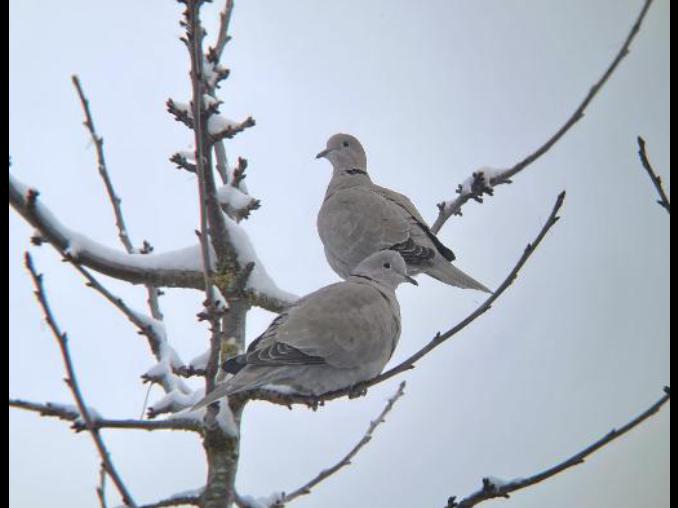} &
        \includegraphics[width=0.22\textwidth]{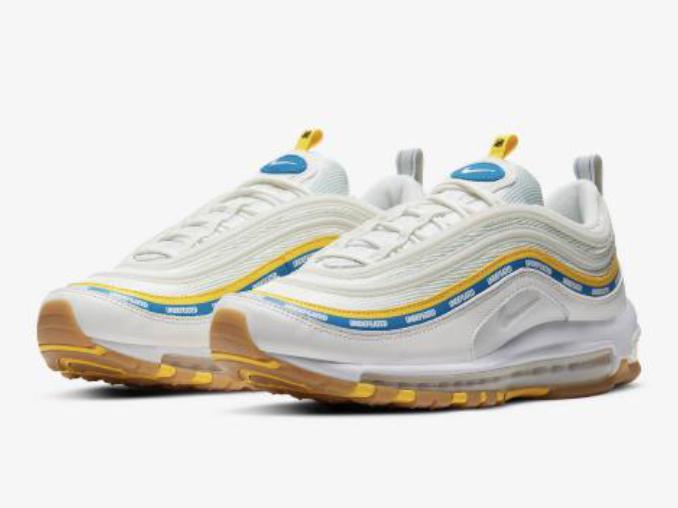} &
        \includegraphics[width=0.22\textwidth]{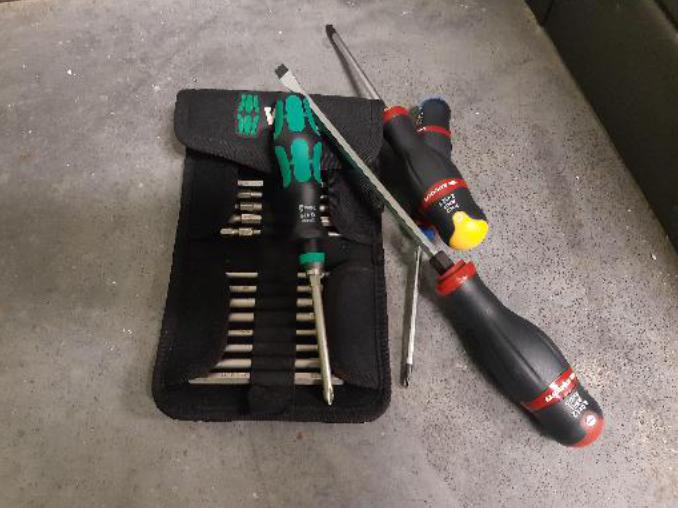} &
        \includegraphics[width=0.22\textwidth]{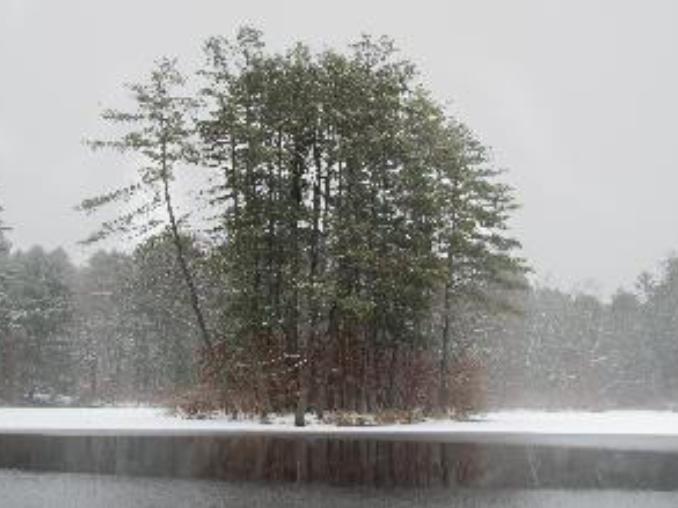} &
        \\
        \blueband \gcc{} &
        person - a gray bird sitting on a branch &
        alternative images of this product &
        the wires are now mounted on the wall. &
        a beautiful white water fountain in the mist &
        \\
        \greenband \dsetname{} &
        \ttbf{r/itookapicture:} \ul{itap of some pigeons} &
        \ttbf{r/sneakers:} \ul{thoughts on these?} &
        \ttbf{r/diy:} \ul{diy tool bag} &
        \ttbf{r/pics:} \ul{my first time seeing snow} &
        \\
        \\
        ~ &
        \includegraphics[width=0.22\textwidth]{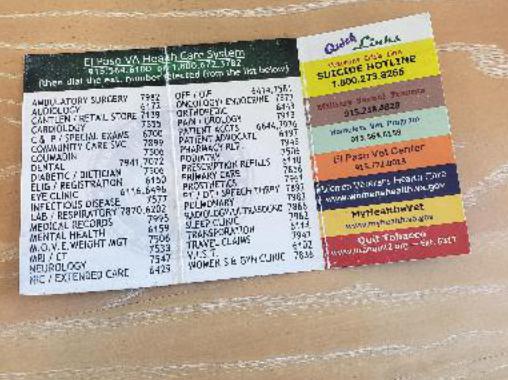} &
        \includegraphics[width=0.22\textwidth]{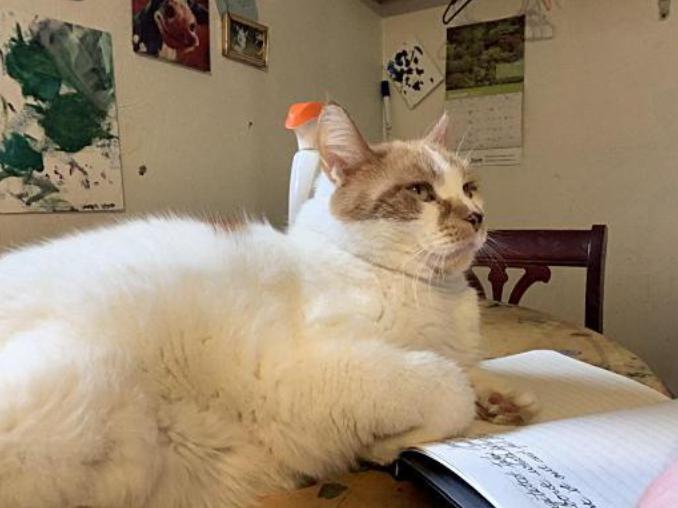} &
        \includegraphics[width=0.22\textwidth]{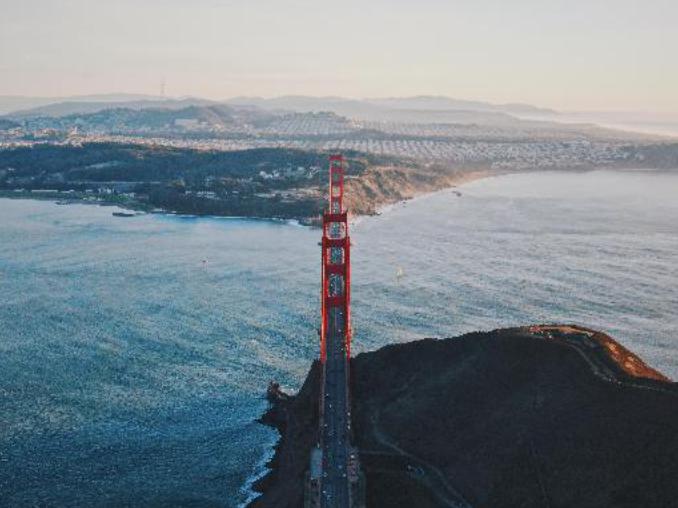} &
        \includegraphics[width=0.22\textwidth]{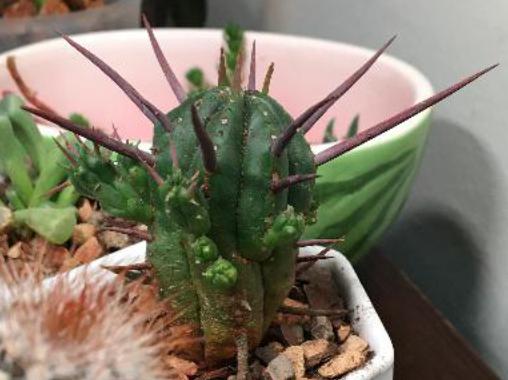} &
        \\
        \blueband \gcc{} &
        \ul{this ticket is not only for sale.} &
        this is what cats look like. &
        the tallest building complex , is currently under construction . &
        \ul{this is a beautiful green cactus plant.} &
        \\
        \greenband \dsetname{} &
        \ttbf{r/mechanicalkeyboards:} i'm not sure if i'm doing this left &
        \ttbf{r/cats:} \ul{my cat is helping me study} &
        \ttbf{r/pics:} \ul{golden gate bridge} &
        \ttbf{r/succulents:} what is this? &
        \\
        \end{tabularx}
    \caption{
        \textbf{Human evaluation: \gcc{} vs. \dsetname{}.}
        This figure includes more examples like \Cref{fig:human_eval} -- 
        \emph{randomly selected} caption predictions from \virtexii{} models pre-trained on \gcc{} and \dsetname{}.
        The underlined caption was chosen by at least two out of three crowd workers, guessed as the human-written caption.
        \dsetname{} predictions are preferred 63.6\% of the time.
    }
    \vspace{-5pt}
    \label{fig:supp_human_eval}
\end{figure}

\clearpage

\section{Subreddit-controlled caption style}
\begin{figure}[h]
    \footnotesize
    \setlength \tabcolsep{3pt}
    \renewcommand{\arraystretch}{1.5}

    \newcommand{\brownband}{\rowcolor{Brown!10}}
    \newcommand{\blueband}{\rowcolor{RoyalBlue!10}}
    \newcommand{\greenband}{\rowcolor{ForestGreen!10}}
    \newcommand{\pinkband}{\rowcolor{RubineRed!10}}
    \newcolumntype{Y}{>{\centering\arraybackslash}X}
    \begin{tabularx}{\linewidth}{YYYYc}
        \includegraphics[width=0.242\textwidth]{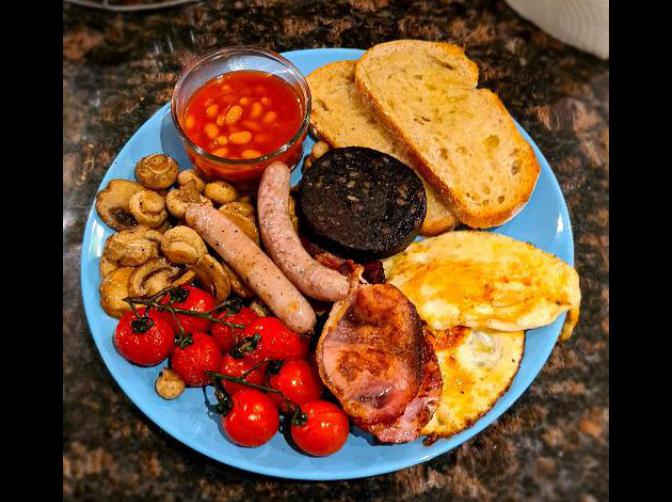} &
        \includegraphics[width=0.242\textwidth]{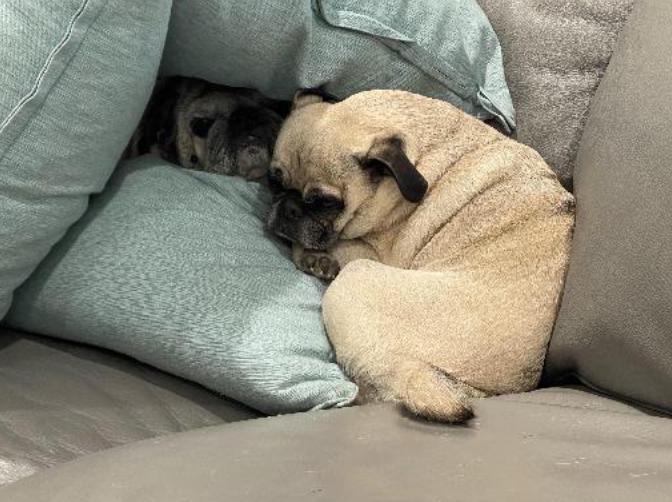} &
        \includegraphics[width=0.242\textwidth]{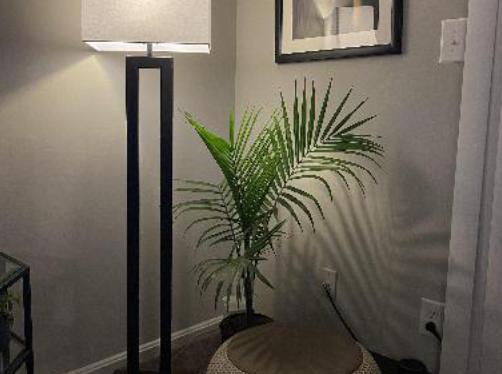} &
        \includegraphics[width=0.242\textwidth]{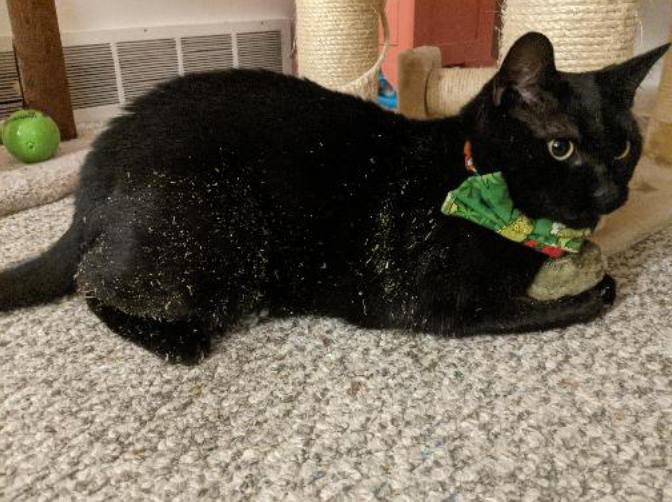} &
        \\
        \brownband     \ttbf{r/food:}  english breakfast &
        \ttbf{r/food:}  i'm not sure if these two are getting ready for dinner tonight. &
        \ttbf{r/food:}  i made a plant stand for my wife's birthday present &
        \ttbf{r/food:}  christmas cat &
        \\
        \blueband     \ttbf{r/thriftstorehauls:}  i found this plate at goodwill for \$5 &
        \ttbf{r/thriftstorehauls:}  found these two pugs in my local thrift store. they are both lonesome and they are so cute. &
        \ttbf{r/thriftstorehauls:}  found this beauty for \$20 &
        \ttbf{r/thriftstorehauls:}  i found a little elf hat for my cat! &
        \\
        \greenband     \ttbf{r/dogpictures:}  my dog ate his breakfast today &
        \ttbf{r/dogpictures:}  my two pugs snuggling under the couch &
        \ttbf{r/dogpictures:}  my dog thinks he's a human &
        \ttbf{r/dogpictures:}  merry christmas from my cat &
        \\
        \pinkband     \ttbf{r/woodworking:}  my first attempt at a full english breakfast &
        \ttbf{r/woodworking:}  i made a pug pillow fort for my dogs. &
        \ttbf{r/woodworking:}  i made a lamp for my wife's birthday present. &
        \ttbf{r/woodworking:}  my cat is very pleased with his christmas present &
        \\
        \\
        \includegraphics[width=0.242\textwidth]{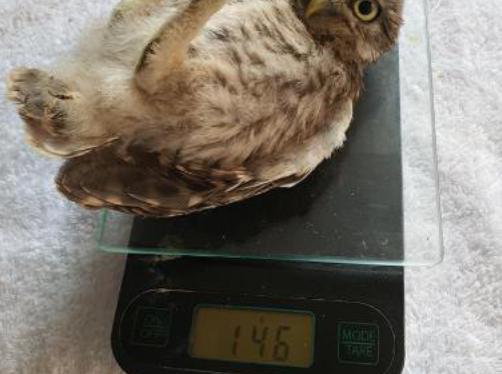} &
        \includegraphics[width=0.242\textwidth]{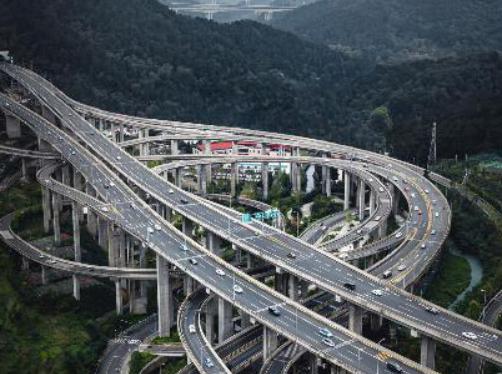} &
        \includegraphics[width=0.242\textwidth]{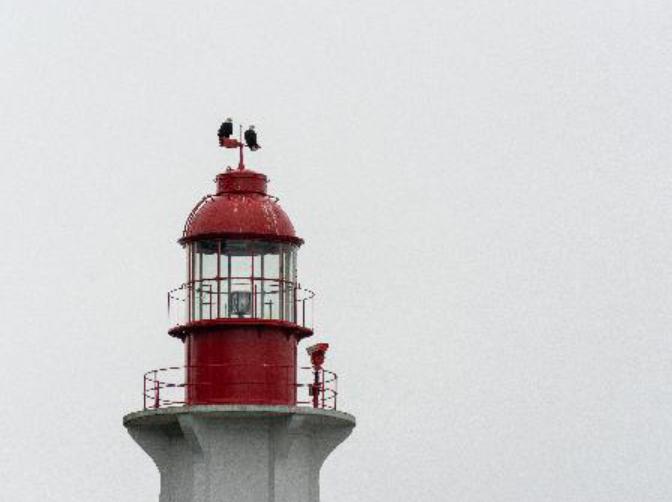} &
        \includegraphics[width=0.242\textwidth]{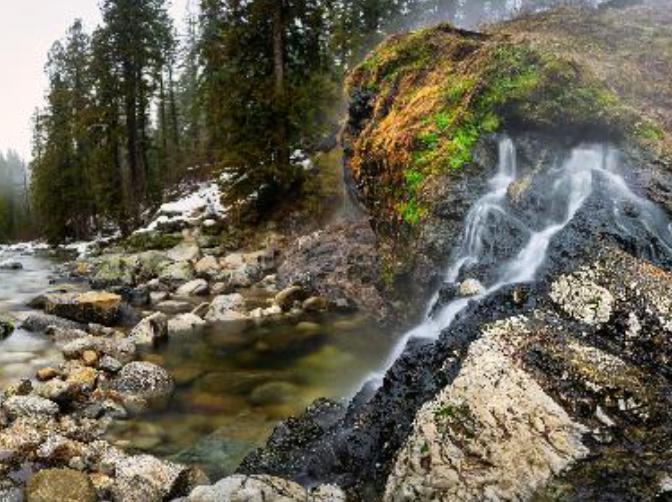} &
        \\
        \brownband     \ttbf{r/amateurphotography:}  i was told you guys would appreciate this. &
        \ttbf{r/amateurphotography:}  i took this picture of a highway interchange in china &
        \ttbf{r/amateurphotography:}  lighthouse &
        \ttbf{r/amateurphotography:}  a waterfall in the rockies &
        \\
        \blueband     \ttbf{r/vintage:}  found this guy in my parents garage. he's been sitting in there for years. &
        \ttbf{r/vintage:}  i've been looking for a few years now. i finally found a bridge in taiwan. &
        \ttbf{r/vintage:}  vintage 2! &
        \ttbf{r/vintage:}  my favorite waterfall &
        \\
        \greenband     \ttbf{r/pics:}  my owl has been in the same spot since i've been working on my phone. &
        \ttbf{r/pics:}  highway interchange between shelbyville and la &
        \ttbf{r/pics:}  lighthouse in the fog &
        \ttbf{r/pics:}  a waterfall in the rockies &
        \\
        \pinkband     \ttbf{r/gardening:}  my garden has been growing a lot of these guys lately. &
        \ttbf{r/gardening:}  i took this picture of a highway interchange in china &
        \ttbf{r/gardening:}  i'm a little bit late but here's my favorite lighthouse &
        \ttbf{r/gardening:}  i'm not sure if this is a good idea but i'm sure. &
        \\
    \end{tabularx}
    \caption{
        \textbf{Subreddit-controlled caption style.}
        This figure includes more examples like Figure 8 of main paper -- 
        \emph{randomly selected} caption predictions from \virtexii{} models pre-trained on \dsetname{}.
        We provide the subreddit names as partial prompts to the model for caption generation,
        for example \ttbf{r/somethingimade} - \inlinecap{[SOS] something i made [SEP]}, and further generate the caption.
    }
    \vspace{-5pt}
    \label{fig:caption_style_supp1}
\end{figure}

\clearpage

\section{Distribution of visual concepts across subreddits}

Each instance in \dsetname{} belongs to one of \dsetsubs{} subreddits.
These subreddits serve as \emph{image labels}, and can cluter visually similar images together.
Here we observe this effect by visualizing the visual feature space of \dsetname{} images per subreddit.

We choose an off-the-shelf ResNeXt-101 32$\times$8d pre-trained on 940M Instagram images~\cite{mahajan2018exploring} as a feature
extractor\footnote{Accessed from \url{https://pytorch.org/hub/facebookresearch_WSL-Images_resnext/}}.
We extract 2048-dimensional global average pooled features for all images and average them per subreddit, resulting in a single 2048-dimensional vector per subreddit.
We perform dimensionality reduction using Barnes Hut T-SNE~\cite{tsne} with default parameters in \texttt{scikit-learn}.

\begin{figure}[h]
  \centering
  \includegraphics[width=\linewidth]{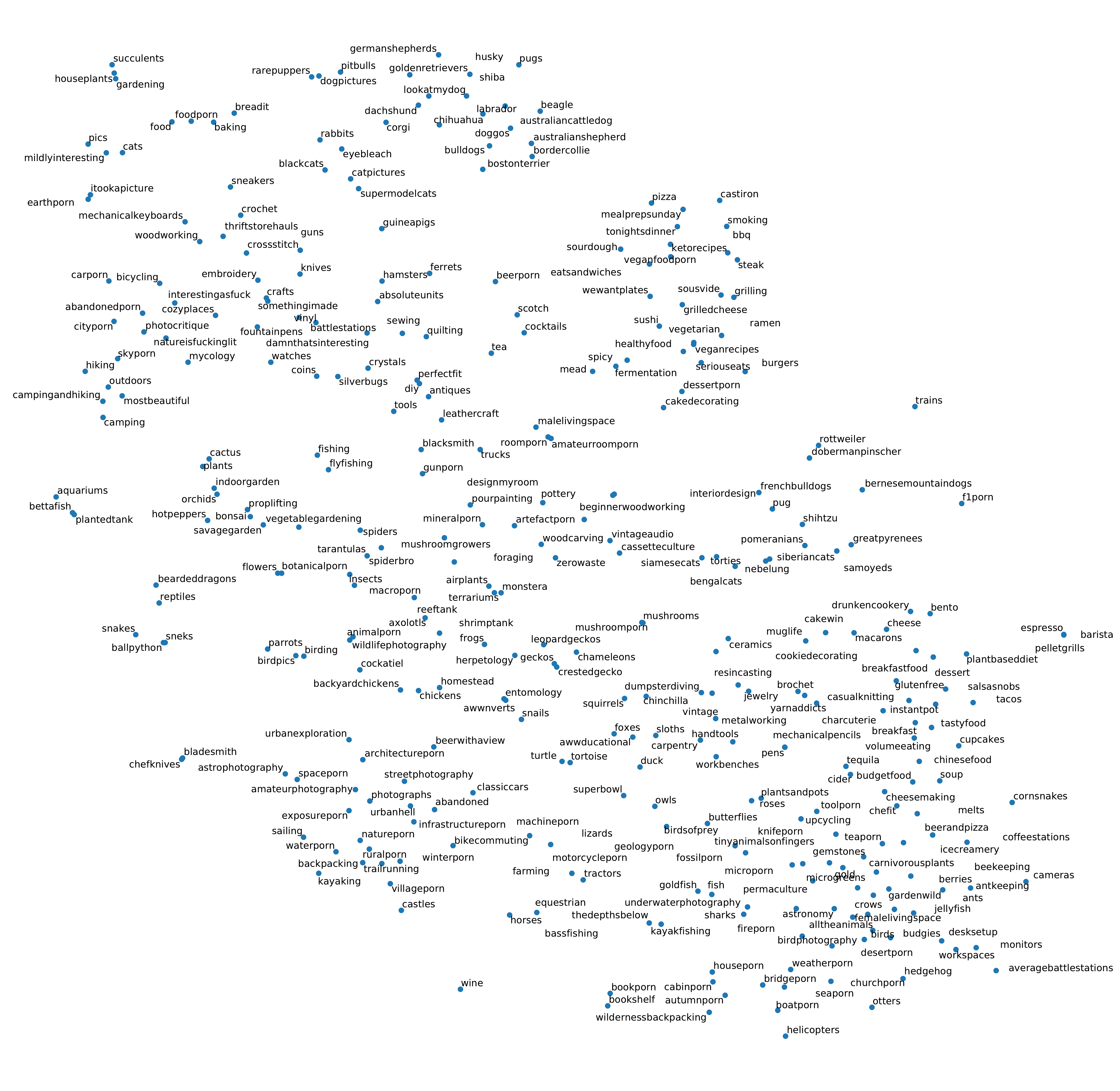}
  \caption{\textbf{T-SNE visualization of image features per subreddit.} Zoom in for better viewing.}
  \label{fig:visual_tsne}
\end{figure}

Visualization is shown below in \Cref{fig:visual_tsne}.
This feature space reveals that subreddits of similar topics form very tight local clusters, such as dogs in top-center (\subreddit{corgi}, \subreddit{husky}, \subreddit{lookatmydog}, \subreddit{pugs}), food and drinks in top-right (\subreddit{bbq}, \subreddit{cocktails}, \subreddit{eatsandwiches}, \subreddit{pizza}, \subreddit{spicy}, \subreddit{tea}). Hence, manually selecting subreddits can let us steer the distribution of visual concepts in \dsetname{}.

\clearpage

\section{Transfer learning experiments: additional details}
\label{appendix:experiments}

\begin{table*}[t]
  \centering
  \small
  \setlength\tabcolsep{2pt}
  \renewcommand{\arraystretch}{1.0}

  \parbox[t]{2mm}{\rotatebox[origin=c]{90}{\textbf{Low-shot \quad Zero Shot\quad\quad\quad\quad}}}
  \hfill
  \begin{tabularx}{0.97\linewidth}{l c XX c XX c XX c XX c XX c XX}
  \toprule
  \multicolumn{1}{l}{\bf \multirow[b]{2}{*}{\shortstack{Pre-train\\Dataset}}}
  && \multicolumn{2}{c}{Pets}
  && \multicolumn{2}{c}{Food}
  && \multicolumn{2}{c}{Flowers}
  && \multicolumn{2}{c}{Cars}
  && \multicolumn{2}{c}{SUN}
  && \multicolumn{2}{c}{Birdsnap}
  \\
  \cmidrule{3-4}
  \cmidrule{6-7}
  \cmidrule{9-10}
  \cmidrule{12-13}
  \cmidrule{15-16}
  \cmidrule{18-19}
  && Top1 & Top5
  && Top1 & Top5
  && Top1 & Top5
  && Top1 & Top5
  && Top1 & Top5
  && Top1 & Top5
  \\
  \midrule
  \sbu{}
  &&   8.7 &  28.8  
  &&   3.0 &  9.8  
  &&  13.7 &  21.5  
  &&   0.6 &   3.3  
  &&  14.7 & 29.6  
  &&   1.3 &   4.7  
  \\
  \gcc{}
  &&  15.5 &  27.4  
  &&  10.9 &  22.1  
  &&  10.1 &  21.2  
  &&   0.5 &   2.8  
  &&  33.3 &  51.3  
  &&   1.6 &   4.6  
  \\
  \band \dsetname-20
  &&  41.8 &  66.2  
  &&  54.6 &  81.8  
  &&  33.5 &  50.9  
  &&   3.2 &  10.1  
  &&  23.9 &  39.3  
  &&  11.8 &  26.0  
  \\
  \band \dsetname{}
  &&  42.4 &  80.2  
  &&  53.8 &  84.0  
  &&  26.2 &  43.6  
  &&   3.1 &  10.8  
  &&  26.8 &  43.4  
  &&   8.3 &  22.1  
  \\
  \midrule
  \sbu{}
  &&  62.0 &  91.1  
  &&  23.3 &  49.2  
  &&  81.0 &  95.2  
  &&   6.7 &  19.4  
  &&  \textbf{19.2} &  \textbf{44.6}  
  &&   1.4 &   6.0  
  \\
  \gcc{}
  &&  63.6 &  91.9  
  &&  17.7 &  41.4  
  &&  74.6 &  90.7  
  &&   4.9 &  15.8  
  &&  17.0 &  41.3  
  &&   1.1 &   4.6  
  \\
  \band \dsetname-20
  &&  72.0 &  \textbf{95.6}  
  &&  \textbf{48.4} &  \textbf{76.8}  
  &&  \textbf{82.7} &  \textbf{95.7}  
  &&  15.6 &  39.1  
  &&  16.8 &  41.4  
  &&   1.5 &   6.1  
  \\
  \band \dsetname{}
  &&  \textbf{73.4} &  95.5  
  &&  45.4 &  74.8  
  &&  80.9 &  95.0  
  &&  \textbf{17.2} &  \textbf{42.3}  
  &&  17.6 &  43.4  
  &&   \textbf{1.6} &   \textbf{6.3}  
  \\
  \bottomrule
\end{tabularx}
\caption{
    \textbf{Additional results: Zero-shot (top-5) and low-shot transfer.}
    We report top-5 accuracy of zero-shot image classification on datasets evaluated in our transfer experiments.
    We also perform low-shot transfer to six datasets -- end-to-end fine-tuning on 1K randomly sampled class-balanced subset of each dataset.
    Models trained on \dsetname{} perform best on all datasets except SUN397.
}
\label{tab:supp_transfer}
\end{table*}

\paragraph{Linear probe image classification:}
We use scikit-learn Logistic Regression with L-BFGS solver, 1000 maximum iterations, and tolerance set to $10^{-4}$.
For each dataset, we hold out a randomly sampled 10\% subset of the training data and use it for validation.
Similar to CLIP, we start with sweeping L2 regularization parameter $\lambda \in \{ 10^{-6}, 10^{-4}, 10^{-2}, 1, 10^{2}, 10^{4}, 10^{6}\}$
and select two $\lambda$ values with highest top-1 accuracy on held out split (these very always consecutive in our experiments).
We \emph{zoom in} the range with eight equally spaced $\lambda$ per decade in logarithmic space to find the best value.
Finally, we use this $\lambda$ to train on the combined training data (including held-out 10\%) and report top-1 accuracy on test split.
The amount of instances in training and test splits used are exactly same as used for evaluating CLIP.

\paragraph{Low-shot classification:}
Another common way of transfer learning is end-to-end finetuning of learned features.
Hence in addition to zero-shot and linear probe classification, here we transfer on low-shot image classification
on a subset of six datasets from the main experiments --
Oxford-IIIT Pets~\cite{parkhi2012pets}, Food-101~\cite{bossard2014food}, Flowers-102~\cite{nilsback2008flowers}, Stanford Cars~\cite{krause2013cars}, SUN-397~\cite{xiao2010sun397}, and Birdsnap~\cite{berg2014birdsnap}.

For each dataset, we randomly sample 1000 instances such that their class distribution stays balanced.
We perform end-to-end fine-tuning of pre-trained weights and follow the same training schedule for every dataset, highly similar to VTAB~\cite{zhai2019large}.
We use SGD with momentum 0.9 and weight decay $10^{-6}$.
We use a batch size of 256 distributed across 8 GPUs (with synchronized BatchNorm~\cite{peng2018megdet}) and train for 5000 iterations ($\sim$1250 epochs with 1K examples).
We use a maximum learning rate of $0.1$ which is multiplied by $0.1$ at iterations 1500, 3000, and 4500.
We use the VISSL~\cite{goyal2021vissl} codebase for all the low-shot transfer experiments.
Results are shown in \Cref{tab:supp_transfer}.
Similar to zero-shot transfer, models trained on \dsetname{} and \dsetname{}-20 perform best on all but one dataset.

  \clearpage
  \section{Datasheet for \dsetname{} dataset}
  \label{appendix:datasheet}
  Datasheets for datasets introduced by \citet{gebru2018datasheets} serve as a medium of communication between the creators and concumers (users) of a dataset.
They effectively consolidate the motivation, creation process, composition, and intended uses of a dataset as a series questions and answers.
In this document, we provide a datasheet for the \dsetname{} dataset.
It accompanies the first version (v1.0) released in October 2021 with our accepted paper at the \emph{NeurIPS 2021 Track on Datasets and Benchmarks}. For the rest of this document:

\begin{compactitem}[\hspace{1pt}-]
    \item All mentions of \emph{\dsetname{}} and all reported data statistics refer to \dsetname{} \texttt{v1.0}.
    \item All mentions of \emph{dataset website} refer to \url{https://redcaps.xyz}.
    \item All mentions of \emph{data collection code} refer to the \texttt{redcaps-downloader} repository available at \url{https://github.com/redcaps-dataset/redcaps-downloader} (also linked on the website).
\end{compactitem}

\section*{Motivation}

\begin{compactenum}[\hspace{0pt}Q1.]
\setcounter{enumi}{0}

\dsquestionex{For what purpose was the dataset created?}{Was there a specific task in mind? Was there a specific gap that needed to be filled? Please provide a description.}
\label{Q1}

\dsanswer{
    Large datasets of image-text pairs are widely used for pre-training generic representations that transfer to a variety of downstream vision and vision-and-language tasks.
    Existing public datasets of this kind were curated from search engine results (SBU Captions~\cite{ordonez2011im2text}) or HTML alt-text from arbitrary web pages (Conceptual Captions~\cite{sharma2018conceptual,changpinyo2021conceptual}).
    They performed complex data filtering to deal with noisy web data.
    Due to aggressive filtering, their data collection is inefficient and diversity is artificially supressed.
    We argue that the quality of data depends on its \emph{source}, and the \emph{human intent} behind its creation.
    In this work, we explore Reddit -- a social media platform, for curating high quality data.
    We introduce \dsetname{} -- a large dataset of \dsetsize{} image-text pairs from Reddit.
    While we expect the use-cases of \dsetname{} to be similar to existing datasets, we discuss how Reddit as a data source leads to fast and lightweight collection, better data quality, lets us easily steer the data distribution, and facilitates ethically responsible data curation.
}

\dsquestion{Who created this dataset (e.g., which team, research group) and on behalf of which entity (e.g., company, institution, organization)?}
\label{Q2}

\dsanswer{
    Four researchers at the University of Michigan (affiliated as of 2021) have created \dsetname{}: Karan Desai, Gaurav Kaul, Zubin Aysola, and Justin Johnson.
}

\dsquestionex{Who funded the creation of the dataset?}{If there is an associated grant, please provide the name of the grantor and the grant name and number.}
\label{Q3}

\dsanswer{
    We collected \dsetname{} without any monetary costs, since no part of our dataset requires annotations from crowd workers or contractors.
    This research work was partially supported by the Toyota Research Institute (TRI).
    However, note that this article solely reflects the opinions and conclusions of its authors and not TRI or any other Toyota entity.
}

\dsquestion{Any other comments?}
\label{Q4}

\dsanswer{No.}

\end{compactenum}


\section*{Composition}

\begin{compactenum}[\hspace{0pt}Q1.]
\setcounter{enumi}{4}
    
\dsquestionex{What do the instances that comprise the dataset represent (e.g., documents, photos, people, countries)?}{ Are there multiple types of instances (e.g., movies, users, and ratings; people and interactions between them; nodes and edges)? Please provide a description.}
\label{Q5}

\dsanswer{
    Each instance in \dsetname{} represents a single Reddit image post.
}

\dsquestion{How many instances are there in total (of each type, if appropriate)?}
\label{Q6}

\dsanswer{
    There are nearly \dsetsize{} (12,011,111) instances in \dsetname{}.
}

\dsquestionex{Does the dataset contain all possible instances or is it a sample (not necessarily random) of instances from a larger set?}{ If the dataset is a sample, then what is the larger set? Is the sample representative of the larger set (e.g., geographic coverage)? If so, please describe how this representativeness was validated/verified. If it is not representative of the larger set, please describe why not (e.g., to cover a more diverse range of instances, because instances were withheld or unavailable).}
\label{Q7}

\dsanswer{
    \dsetname{} is a small sample drawn from all the data uploaded to Reddit.
    Millions of Reddit users submit image posts across thousands of subreddits on a daily basis.
    We hand-picked \dsetsubs{} subreddits containing high-quality photographs with descriptive captions, while leaving out lots of subreddits focused on many other topics like politics, religion, science, and memes.
    Even within the selected subreddits, we filtered instances to improve data quality and mitigate privacy risks for people appearing images.
    Hence, \dsetname{} data does not fully represent Reddit.
}

\dsquestionex{What data does each instance consist of? “Raw” data (e.g., unprocessed text or images) or features?}{In either case, please provide a description.}
\label{Q8}

\dsanswer{
    Each instance in \dsetname{} consists of nine metadata fields:
    \begin{compactitem}
        \item \texttt{"image\_id"}: Unique alphanumeric ID of the image post (assigned by Reddit).
        \item \texttt{"author"}: Reddit username of the image post author.
        \item \texttt{"url"}: Static URL for downloading the image associated with the post.
        \item \texttt{"raw\_caption"}: Textual description of the image, written by the post author.
        \item \texttt{"caption"}: Cleaned version of \texttt{"raw\_caption"} by us (see \qref{Q35}).
        \item \texttt{"subreddit"}: Name of subreddit where the post was submitted.
        \item \texttt{"score"}: Net upvotes (discounting downvotes) received by the image post.
        \item \texttt{"created\_utc"}: Integer time epoch (in UTC) when the post was submitted to Reddit.
        \item \texttt{"permalink"}: Partial URL of the Reddit post (\texttt{https://reddit.com/<permalink>}).
    \end{compactitem}
}

\dsquestionex{Is there a label or target associated with each instance?}{If so, please provide a description.}
\label{Q9}

\dsanswer{
    No, we do not define any label or target for the instances.
    Targets are task-dependent.
    \dsetname{} can be used for a variety of tasks such as
    image captioning (\textit{inputs = images, targets = captions}),
    image classification (\textit{inputs = images, targets = subreddits}),
    text-to-image generation (\textit{inputs = captions, targets = images}),
    or self-supervised visual learning (\textit{inputs = images, no targets}).
}

\dsquestionex{Is any information missing from individual instances?}{If so, please provide a description, explaining why this information is missing (e.g., because it was unavailable). This does not include intentionally removed information, but might include, e.g., redacted text.}
\label{Q10}

\dsanswer{
    No and yes.
    No, because all the metadata fields for every instance are filled with valid values.
    Yes, because the \texttt{"url"} for some instances may not retrieve the underlying image.
    This may happen if the Reddit user (author) removes the post from Reddit.
    Such deletions reduce our dataset size over time, however post deletions are very rare after six months of creation.
}

\dsquestionex{Are relationships between individual instances made explicit (e.g., users’ movie ratings, social network links)?}{If so, please describe how these relationships are made explicit.}
\label{Q11}

\dsanswer{
    Some implicit relationships do exist in our data.
    All instances belonging to the same subreddit are likely to have high related visual and textual content.
    Moreover, multiple images posted by a single Reddit user may be highly related (photos of their pets, cars, etc.).
}

\dsquestionex{Are there recommended data splits (e.g., training, development/validation, testing)?}{If so, please provide a description of these splits, explaining the rationale behind them.}
\label{Q12}

\dsanswer{
    We intend our dataset to be primarily used for pre-training with one or more specific downstream task(s) in mind.
    Hence, all instances in our dataset would be used for training while the validation split is derived from downstream task(s).
    If users require a validation split, we recommend sampling it such that it follows the same subreddit distribution as entire dataset.
}

\dsquestionex{Are there any errors, sources of noise, or redundancies in the dataset?}{If so, please provide a description.}
\label{Q13}

\dsanswer{
    \dsetname{} is noisy \emph{by design} since image-text pairs on the internet are noisy and unstructured.
    Some instances may also have duplicate images and captions --
    Reddit users may have shared the same image post in multiple subreddits.
    Such redundancies constitute a very small fraction of the dataset, and should have almost no effect in training large-scale models.
}

\dsquestionex{Is the dataset self-contained, or does it link to or otherwise rely on external resources (e.g., websites, tweets, other datasets)?}{If it links to or relies on external resources,
\begin{compactenum}[\hspace{1pt}(a)]
    \item Are there guarantees that they will exist, and remain constant, over time?
    \item Are there official archival versions of the complete dataset (i.e., including the external resources as they existed at the time the dataset was created)?
    \item Are there any restrictions (e.g., licenses, fees) associated with any of the external resources that might apply to a future user? Please provide descriptions of all external resources and any restrictions associated with them, as well as links or other access points, as appropriate.
\end{compactenum}}
\label{Q14}

\dsanswer{
    We do not distribute images of our dataset to respect Reddit user privacy and to limit our storage budget.
    Instead we provide image URLs (\texttt{"url"}, \qref{Q8}) that point to images hosted on either Reddit, Imgur, or Flickr image servers.
    In response to sub-questions:
    \begin{compactenum}[\hspace{1pt}(a)]
        \item These image servers ensure stable access unless the Reddit user deletes their image post.
        \item Yes, Reddit archives all the metadata of submitted posts. For images, Reddit only archives the URL and not the media content, giving full control of accessibility to the users.
        \item All image URLs are freely accessible. It is unlikely for the image servers to restrict access in the future, given their free accessibility over the past decade.
    \end{compactenum}
}

\dsquestionex{Does the dataset contain data that might be considered confidential (e.g., data that is protected by legal privilege or by doctor-patient confidentiality, data that includes the content of individuals non-public communications)?}{If so, please provide a description.}
\label{Q15}

\dsanswer{
    No, the subreddits included in \dsetname{} do not cover topics that may be considered confidential.
    All posts were publicly shared on Reddit prior to inclusion in \dsetname{}.
}

\dsquestionex{Does the dataset contain data that, if viewed directly, might be offensive, insulting, threatening, or might otherwise cause anxiety?}{If so, please describe why.}
\label{Q16}

\dsanswer{
    The scale of \dsetname{} means that we are unable to verify the contents of all images and captions.
    However we have tried to minimize the possibility that \dsetname{} contains data that might be offensive, insulting, threatening, or might cause anxiety via the following mitigations:
    \begin{compactenum}
        \item We manually curate the set of subreddits from which to collect data; we only chose subreddits that are not marked NSFW and which generally contain non-offensive content.
        \item Within our curated subreddits, we did not include any posts marked NSFW.
        \item We removed all instances whose captions contained any of the 400 potentially offensive words or phrases\footnote{\url{https://github.com/LDNOOBW/List-of-Dirty-Naughty-Obscene-and-Otherwise-Bad-Words}}. Refer \textcolor{Red}{Section 2.2} in the main paper.
        \item We remove all instances whose images were flagged NSFW by an off-the-shelf detector. We manually checked 50K random images in \dsetname{} and found one image containing nudity (exposed buttocks; no identifiable face). Refer \textcolor{Red}{Section 2.2} in the main paper.
    \end{compactenum}
}

\dsquestionex{Does the dataset relate to people?}{If not, you may skip remaining questions in this section.}
\label{Q17}

\dsanswer{
    The dataset pertains to people in that people wrote the captions and posted images to Reddit that we curate in \dsetname{}.
    We made specific design choices while curating \dsetname{} to avoid large quantities of images containing people:
    \begin{compactenum}
        \item We collect data from manually curated subreddits in which most contain primarily pertains to animals, objects, places, or activities.
        We exclude all subreddits whose primary purpose is to share and describe images of people (such as celebrity photos or user selfies).
        \item We use an off-the-shelf face detector to find and remove images with potential presence of human faces.
        We manually checked 50K random images in \dsetname{} (\qref{Q16}) and found 79 images with identifiable human faces -- the entire dataset may have $\approx$19K (0.15\%) images with identifiable people.
        Refer \textcolor{Red}{Section 2.2} in the main paper.
    \end{compactenum}
}

\dsquestionex{Does the dataset identify any subpopulations (e.g., by age, gender)?}{If so, please describe how these subpopulations are identified and provide a description of their respective distributions within the dataset.}
\label{Q18}

\dsanswer{
    \dsetname{} does not explicitly identify any subpopulations.
    Since some images contain people and captions are free-form natural language written by Reddit users, it is possible that some captions may identify people appearing in individual images as part of a subpopulation.
}

\dsquestionex{Is it possible to identify one or more natural persons, either directly or indirectly (i.e., in combination with other data) from the dataset?}{If so, please describe how.}
\label{Q19}

\dsanswer{
    Yes, all instances in \dsetname{} include Reddit usernames of their post authors.
    This could be used to look up the Reddit user profile, and some Reddit users may have identifying information in their profiles.
    Some images may contain human faces (\qref{Q17}) which could be identified by appearance.
    However, note that all this information is already public on Reddit, and searching it in \dsetname{} is no easier than searching directly on Reddit.
}

\dsquestionex{Does the dataset contain data that might be considered sensitive in any way (e.g., data that reveals racial or ethnic origins, sexual orientations, religious beliefs, political opinions or union memberships, or locations; financial or health data; biometric or genetic data; forms of government identification, such as social security numbers; criminal history)?}{If so, please provide a description.}
\label{Q20}

\dsanswer{
    Highly unlikely, the data from our manually selected subreddits does not contain sensitive information of the above forms.
    In case some instances have such information, then note that all this information is already publicly available on Reddit.
}

\dsquestion{Any other comments?}
\label{Q21}

\dsanswer{No.}

\end{compactenum}


\section*{Collection Process}

\begin{compactenum}[\hspace{0pt}Q1.]
\setcounter{enumi}{21}

\dsquestionex{How was the data associated with each instance acquired?}{Was the data directly observable (e.g., raw text, movie ratings), reported by subjects (e.g., survey responses), or indirectly inferred/derived from other data (e.g., part-of-speech tags, model-based guesses for age or language)? If data was reported by subjects or indirectly inferred/derived from other data, was the data validated/verified? If so, please describe how.}
\label{Q22}

\dsanswer{
    We collected instance IDs using Pushshift API (\url{https://pushshift.io}) and remaining metadata fields (\qref{Q8}) using the Reddit API (\url{https://www.reddit.com/wiki/api}).
    All fields except \texttt{"caption"} are available in API responses; \texttt{"caption"} is derived by applying text pre-processing to \texttt{"raw\_caption"} field (\qref{Q35}).
}

\dsquestionex{What mechanisms or procedures were used to collect the data (e.g., hardware apparatus or sensor, manual human curation, software program, software API)?}{How were these mechanisms or procedures validated?}
\label{Q23}

\dsanswer{
    We collected all data using compute resources at the University of Michigan.
    The code for querying APIs and filtering data is implemented in Python.
    We validated our implementation by manually checking few \dsetname{} instances with their posts on \url{https://reddit.com}.
}

\dsquestion{If the dataset is a sample from a larger set, what was the sampling strategy?}
\label{Q24}

\dsanswer{
    \dsetname{} is a small sample containing data from \dsetsubs{} subreddits out of thousands of subreddits on Reddit.
    We hand-picked each subreddit for our dataset based on its content.
    See \qref{Q7}, \qref{Q16}, and \qref{Q17} for details on how we selected each subreddit.
}

\dsquestion{Who was involved in data collection process (e.g., students, crowd-workers, contractors) and how were they compensated (e.g., how much were crowd-workers paid)?}
\label{Q25}

\dsanswer{
    Our data collection pipeline is fully automatic and does not require any human annotators.
    Reddit users have uploaded image posts whose metadata is a part of \dsetname{} -- we did not directly interact with these users.
}

\dsquestionex{Over what timeframe was the data collected? Does this timeframe match the creation timeframe of the data associated with the instances (e.g., recent crawl of old news articles)?}{If not, please provide a description of the timeframe.}
\label{Q26}

\dsanswer{
    \dsetname{} contains image posts that were uploaded to Reddit between 2008--2020.
    We collected all data in early 2021, which we used to conduct experiments for our NeurIPS 2021 submission.
    Since Reddit posts may get deleted over time, we exactly re-collected a fresh version in August 2021 after acceptance (and re-trained all our experiments).
    Reddit posts observe the most user activity (upvotes, comments, moderation) for six months after their creation -- posts from 2008--2020 are less likely to be updated after August 2021.
}

\dsquestionex{Were any ethical review processes conducted (e.g., by an institutional review board)?}{If so, please provide a description of these review processes, including the outcomes, as well as a link or other access point to any supporting documentation.}
\label{Q27}

\dsanswer{
    We did not conduct a formal ethical review process via institutional review boards.
    However, as described in \textcolor{Red}{Section 2.2} of the main paper and \qref{Q16} we employed several filtering mechanisms to try and remove instances that could be problematic.
}

\dsquestionex{Does the dataset relate to people?}{If not, you may skip remaining questions in this section.}
\label{Q28}

\dsanswer{
    Some images of \dsetname{} may contain images of people (see \qref{Q17}).
}

\dsquestion{Did you collect the data from the individuals in question directly, or obtain it via third parties or other sources (e.g., websites)?}
\label{Q29}

\dsanswer{
    We collected data submitted by Reddit users indirectly through the Reddit API.
    However, users agree with Reddit's User Agreement regarding redistribution of their data by Reddit.
}

\dsquestionex{Were the individuals in question notified about the data collection?}{If so, please describe (or show with screenshots or other information) how notice was provided, and provide a link or other access point to, or otherwise reproduce, the exact language of the notification itself.}
\label{Q30}

\dsanswer{
    No. Reddit users are anonymous by default, and are not required to share their personal contact information (email, phone numbers, etc.).
    Hence, the only way to notify the authors of \dsetname{} image posts is by sending them private messages on Reddit.
    This is practically difficult to do manually, and will be classified as spam and blocked by Reddit if attempted to programmatically send a templated message to millions of users.
}

\dsquestionex{Did the individuals in question consent to the collection and use of their data?}{If so, please describe (or show with screenshots or other information) how consent was requested and provided, and provide a link or other access point to, or otherwise reproduce, the exact language to which the individuals consented.}
\label{Q31}

\dsanswer{
    Users did not explicitly consent to the use of their data in our dataset.
    However, by uploading their data on Reddit, they consent that it would appear on the Reddit plaform and will be accessible via the official Reddit API (which we use to collect \dsetname{}).
}

\dsquestionex{If consent was obtained, were the consenting individuals provided with a mechanism to revoke their consent in the future or for certain uses?}{If so, please provide a description, as well as a link or other access point to the mechanism (if appropriate).}
\label{Q32}

\dsanswer{
    Users have full control over the presence of their data in our dataset.
    If users wish to revoke their consent, they can delete the underlying Reddit post -- it will be automatically removed dfrom \dsetname{} since we distributed images as URLs.
    Moreover, we provide an opt-out request form on our dataset website for anybody to request removal of an individual instance if it is potentially harmful (e.g. NSFW, violates privacy, harmful stereotypes, etc.).
}

\dsquestionex{Has an analysis of the potential impact of the dataset and its use on data subjects (e.g., a data protection impact analysis) been conducted?}{If so, please provide a description of this analysis, including the outcomes, as well as a link or other access point to any supporting documentation.}
\label{Q33}

\dsanswer{No.}

\dsquestion{Any other comments?}
\label{Q34}

\dsanswer{No.}

\end{compactenum}


\section*{Preprocessing, Cleaning, and/or Labeling}

\begin{compactenum}[\hspace{0pt}Q1.]
\setcounter{enumi}{34}

\dsquestionex{Was any preprocessing/cleaning/labeling of the data done (e.g., discretization or bucketing, tokenization, part-of-speech tagging, SIFT feature extraction, removal of instances, processing of missing values)?}{If so, please provide a description. If not, you may skip the remainder of the questions in this section.}
\label{Q35}

\dsanswer{
    We filtered all image posts with $< 2$ net upvotes, and those marked NSFW on Reddit.
    We remove character accents, emojis, non-latin characters, sub-strings enclosed in brackets (\inlinecap{(.*)}, \inlinecap{[.*]}), and replace social media handles (words starting with `@') with a special \inlinecap{[USR]} token.
    Refer \textcolor{Red}{Section 2.1} in the main paper for more details.
    We also remove additional instances with focus on ethical considerations, see 
    \qref{Q16}, \qref{Q17} for more details.
}

\dsquestionex{Was the ``raw'' data saved in addition to the preprocessed/cleaned/labeled data (e.g., to support unanticipated future uses)?}{If so, please provide a link or other access point to the “raw” data.}
\label{Q36}

\dsanswer{
    We provide the unprocessed captions obtained as-is from Reddit as part of our annotations (see \texttt{``raw\_caption''} in \qref{Q8}).
    However, we entirely discard all instances that were filtered with ethical considerations -- based on presence of faces, NSFW content, or harmful language.
}

\dsquestionex{Is the software used to preprocess/clean/label the instances available?}{If so, please provide a link or other access point.}
\label{Q37}

\dsanswer{
    Yes, the data collection code is open-sourced and accessible from the dataset website.
}

\dsquestion{Any other comments?}
\label{Q38}

\dsanswer{No.}

\end{compactenum}


\section*{Uses}

\begin{compactenum}[\hspace{0pt}Q1.]
\setcounter{enumi}{38}

\dsquestionex{Has the dataset been used for any tasks already?}{If so, please provide a description.}
\label{Q39}

\dsanswer{
    We have used our dataset to train deep neural networks that perform image captioning, and that learn transferable visual representations for a variety of downstream visual recognition tasks (image classification, object detection, instance segmentation).
}

\dsquestionex{Is there a repository that links to any or all papers or systems that use the dataset?}{If so, please provide a link or other access point.}
\label{Q40}

\dsanswer{
    We do not maintain such a repository.
    However, citation trackers like Google Scholar and Semantic Scholar would list all future works that cite our dataset.
}

\dsquestion{What (other) tasks could the dataset be used for?}
\label{Q41}

\dsanswer{
    We anticipate that the dataset could be used for a variety of vision-and-language (\vnl{}) tasks, such as image or text retrieval or text-to-image synthesis.
}

\dsquestionex{Is there anything about the composition of the dataset or the way it was collected and preprocessed/cleaned/labeled that might impact future uses?}{For example, is there anything that a future user might need to know to avoid uses that could result in unfair treatment of individuals or groups (e.g., stereotyping, quality of service issues) or other undesirable harms (e.g., financial harms, legal risks) If so, please provide a description. Is there anything a future user could do to mitigate these undesirable harms?}
\label{Q42}

\dsanswer{
    This is very difficult to anticipate.
    Future users of our dataset should be aware of Reddit's user demographics (as described in \textcolor{Red}{Section 2.2} of the main paper) which might subtly influence the types of images, languages, and ideas that are present in the dataset.
    Moreover, users should be aware that our dataset intentionally excludes data from subreddits whose primary purpose is to share images that depict or describe people.
}

\dsquestionex{Are there any tasks for which the dataset should not be used?}{If so, please provide a description.}
\label{Q43}

\dsanswer{
    Broadly speaking, our dataset should only be used for non-commercial academic research.
    Our dataset should not be used for any tasks that involve identifying features related to people (facial recognition, gender, age, ethnicity identification, etc.) or make decisions that impact people (mortgages, job applications, criminal sentences; or moderation decisions about user-uploaded data that could result in bans from a website).
    Any commercial and for-profit uses of our dataset are restricted -- it should not be used to train models that will be deployed in production systems as part of a product offered by businesses or government agencies.
}

\dsquestion{Any other comments?}
\label{Q44}

\dsanswer{No.}

\end{compactenum}


\section*{Distribution}

\begin{compactenum}[\hspace{0pt}Q1.]
\setcounter{enumi}{44}

\dsquestionex{Will the dataset be distributed to third parties outside of the entity (e.g., company, institution, organization) on behalf of which the dataset was created?}{If so, please provide a description.}
\label{Q45}

\dsanswer{
    Yes, our dataset will be publicly available.
}

\dsquestionex{How will the dataset will be distributed (e.g., tarball on website, API, GitHub)}{Does the dataset have a digital object identifier (DOI)?}
\label{Q46}

\dsanswer{
    We distribute our dataset as a ZIP file containing all the annotations (JSON files).
    Users will have to download the images by themselves by using our data collection code.
    All uses of \dsetname{} should cite the NeurIPS 2021 paper as the reference.
}

\dsquestion{When will the dataset be distributed?}
\label{Q47}

\dsanswer{
    The dataset will be publicly available starting from October 2021.
}

\dsquestionex{Will the dataset be distributed under a copyright or other intellectual property (IP) license, and/or under applicable terms of use (ToU)?}{If so, please describe this license and/or ToU, and provide a link or other access point to, or otherwise reproduce, any relevant licensing terms or ToU, as well as any fees associated with these restrictions.}
\label{Q48}

\dsanswer{
    Uses of our dataset are subject to Reddit API terms (\url{https://www.reddit.com/wiki/api-terms}).
    Additionally users must comply with Reddit User Agreeement, Content Policy, and Privacy Policy -- all accessible at \url{https://www.redditinc.com/policies}.
    The data collection code is released with an MIT license.
}

\dsquestionex{Have any third parties imposed IP-based or other restrictions on the data associated with the instances?}{If so, please describe these restrictions, and provide a link or other access point to, or otherwise reproduce, any relevant licensing terms, as well as any fees associated with these restrictions.}
\label{Q49}

\dsanswer{
    The images corresponding to our instances are legally owned by Reddit users.
    Our dataset users can download them from the URLs we provide in annotation files, but resdistributing images for commercial use is prohibited.
}

\dsquestionex{Do any export controls or other regulatory restrictions apply to the dataset or to individual instances?}{If so, please describe these restrictions, and provide a link or other access point to, or otherwise reproduce, any supporting documentation.}
\label{Q50}

\dsanswer{No.}

\dsquestion{Any other comments?}
\label{Q51}

\dsanswer{No.}

\end{compactenum}


\section*{Maintenance}

\begin{compactenum}[\hspace{0pt}Q1.]
\setcounter{enumi}{51}

\dsquestion{Who will be supporting/hosting/maintaining the dataset?}
\label{Q52}

\dsanswer{
    The dataset is hosted using Dropbox service provided by the University of Michigan.
    All the information about the dataset, including links to the paper, code, and future announcements will be accessible at the dataset website (\url{https://redcaps.xyz}).
}

\dsquestion{How can the owner/curator/manager of the dataset be contacted (e.g., email address)?}
\label{Q53}

\dsanswer{The contact emails of authors is available on the dataset website and in this datasheet.}

\dsquestionex{Is there an erratum?}{If so, please provide a link or other access point.}
\label{Q54}

\dsanswer{
    There is no erratum for our initial release.
    We will version all errata as future releases (\qref{Q55}) and document them on the dataset website.
}

\dsquestionex{Will the dataset be updated (e.g., to correct labeling errors, add new instances, delete instances)?}{If so, please describe how often, by whom, and how updates will be communicated to users (e.g., mailing list, GitHub)?}
\label{Q55}

\dsanswer{
    We will update our dataset once every year and announce it on the dataset website.
    These future versions would include new instances corresponding to image posts made in 2021 and beyond,
    would remove instances that were requested to be removed via the opt out form (\qref{Q32}).
}

\dsquestionex{If the dataset relates to people, are there applicable limits on the retention of the data associated with the instances (e.g., were individuals in question told that their data would be retained for a fixed period of time and then deleted)?}{If so, please describe these limits and explain how they will be enforced.}
\label{Q56}

\dsanswer{
    Some images in \dsetname{} may depict people (\qref{Q17}).
    Rather then directly distributing images, we distribute URLs that point to the original images uploaded by Reddit users.
    This means that users retain full control of their data -- any post deleted from Reddit will be automatically removed from \dsetname{} (see also \qref{Q10}, \qref{Q14}, \qref{Q31}).
}

\dsquestionex{Will older versions of the dataset continue to be supported/hosted/maintained?}{If so, please describe how. If not, please describe how its obsolescence will be communicated to users.}
\label{Q57}

\dsanswer{
    A new version release of \dsetname{} will automatically deprecate its previous version.
    We will only support and maintain the latest version at all times.
    Deprecated versions will remain accessible on the dataset website for a few weeks, after which they will be removed.
    We decided to deprecate old versions to ensure that any data that is requested to be removed (\qref{Q32}) will be no longer accessible in future versions.
}

\dsquestionex{If others want to extend/augment/build on/contribute to the dataset, is there a mechanism for them to do so?}{If so, please provide a description. Will these contributions be verified? If so, please describe how. If not, why not? Is there a process for communicating/distributing these contributions to other users? If so, please provide a description.}
\label{Q58}

\dsanswer{
    Anyone can extend \dsetname{} by using our data collection code (linked on the website).
    We are open to accept extensions via personal communication with contributors.
    Otherwise, our code and data licenses allow others to create independent derivative works (with proper attribution) as long as they are used for non-commercial academic research.
}
\end{compactenum}

\end{appendices}

{\small
\bibliographystyle{unsrtnat}
\bibliography{references}
}

\end{document}